%% file: main.tex
\documentclass[10pt,twocolumn,letterpaper]{article}

\usepackage{iccv}      
\usepackage{tcolorbox}

\usepackage[accsupp]{axessibility} 

\input{preamble}

\definecolor{iccvblue}{rgb}{0.21,0.49,0.74}
\usepackage[pagebackref,breaklinks,colorlinks,allcolors=iccvblue]{hyperref}

\def\paperID{7506} 
\def\confName{ICCV}
\def\confYear{2025}

\title{Correspondence-Free Fast and Robust Spherical Point Pattern Registration}

\author{Anik Sarker \qquad Alan T. Asbeck \\
Dept. of Mechanical Engineering, Virginia Tech \\
{\tt\small \{aniks, aasbeck\}@vt.edu}
}

\begin{document}
\maketitle
\input{sec/0_abstract}

\input{sec/1_intro}

\input{sec/2_Related_work}
\input{sec/3_Mathematical_preliminaries}

\input{sec/4_Methods}

\input{sec/5_Experiments}

\input{sec/6_Conclussion}

{
    \small
    \bibliographystyle{ieeenat_fullname}
    \bibliography{main}
}

\input{sec/X_suppl}

\end{document}

%% file: preamble.tex
\usepackage{svg}
\usepackage{mathtools}
\usepackage{algorithm}
\usepackage{algpseudocode}
\usepackage{soul}

%% file: sec/0_abstract.tex
\begin{abstract}





Current methods to estimate the rotation between two spherical (\(\mathbb{S}^2\)) patterns typically rely on maximizing their spherical cross-correlation. However, these approaches exhibit computational complexities greater than cubic \(O(n^3)\) with respect to rotation space discretization.
We propose a rotation estimation algorithm between two spherical patterns with linear time complexity \(O(n)\). Unlike existing methods, we explicitly represent spherical patterns as discrete 3D point sets on the unit sphere, reformulating rotation estimation as a spherical point-set alignment (i.e., the Wahba problem for 3D unit vectors). We introduce three novel algorithms: (1) SPMC (Spherical Pattern Matching by Correlation), (2) FRS (Fast Rotation Search), and (3) a hybrid approach (SPMC+FRS) that combines the advantages of the previous two methods. Our experiments demonstrate that in the \(\mathbb{S}^2\) domain and in correspondence-free settings, our algorithms are over 10x faster and over 10x more accurate than current state-of-the-art methods for the Wahba problem with outliers. We validate our approach through extensive simulations on a new dataset of spherical patterns, the ``Robust Vector Alignment Dataset."

Furthermore, we adapt our methods to two real-world tasks: (i) Point Cloud Registration (PCR) and (ii) rotation estimation for spherical images. In the PCR task, our approach successfully registers point clouds exhibiting overlap ratios as low as 65\%. In spherical image alignment, we show that our method robustly estimates rotations even under challenging conditions involving substantial clutter (over 19\%) and large rotational offsets. Our results highlight the effectiveness and robustness of our algorithms in realistic, complex scenarios.
Our dataset and code are available at: \href{https://github.com/ARLab-VT/Robust-Vector-Set-Alignment}{https://github.com/ARLab-VT/Robust-Vector-Set-Alignment}

\end{abstract}

%% file: sec/1_intro.tex
\section{Introduction}
\label{sec:intro}

\begin{figure}[t!]
  \centering
   \includegraphics[width=1\linewidth]{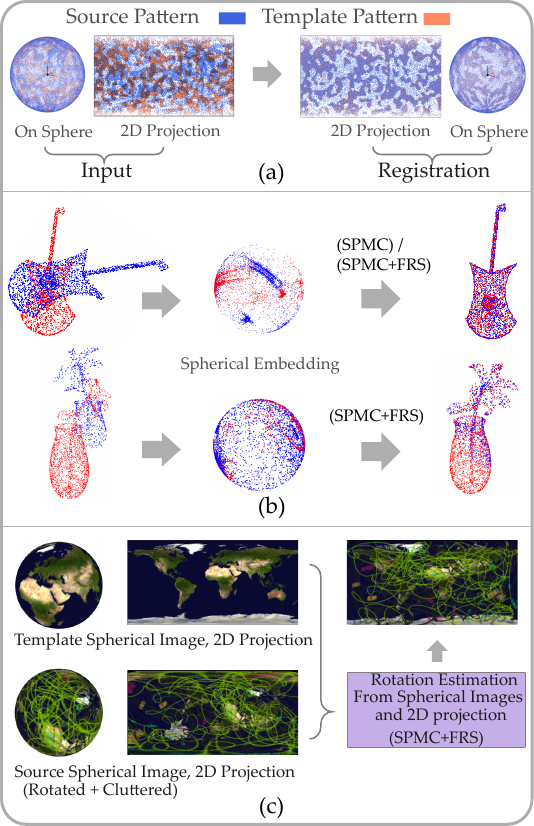}
   \caption{Applications of our algorithms. (a) The spherical point pattern registration algorithm takes a template point cloud and a rotated, noisy source point cloud and performs registration to align the source to the template. (b) 3D object point clouds can be embedded onto a sphere using spherical embeddings, such as the Extended Gaussian Image (EGI) or our method, Centroid Aware Spherical Embedding (CASE), to perform rotation estimation. (c) Our method can be adapted for spherical image registration with large perturbations and noise.}
   
   \label{fig:onecol_overview}
\end{figure}








The classical \emph{Wahba Problem} \cite{wahba1965least} seeks to determine the rotation between two coordinate frames based on vector observations in each frame. Alongside the Wahba problem, the closely related \emph{Orthogonal Procrustes Problem} \cite{gower2004procrustes} is foundational to \emph{rotation search} \cite{bazin2014globally, hartley2009global}. Rotation search is a fundamental task in various applications, including attitude estimation \cite{chin2019star, cheng2019total}, point cloud registration \cite{besl1992method, chetverikov2002trimmed, li20073d}, image stitching \cite{parra2015guaranteed, yang2019quaternion}, 3D reconstruction \cite{blais1995registering}, and robotics \cite{bernreiter2021phaser}, 
among others.
For closed-form cases where true one-to-one correspondences are provided, the problem is considered solved with elegant solutions such as those in \cite{kabsch1976solution, horn1987closed, arun1987least}. However, in real-world, non-closed-form scenarios, the problem remains challenging and far from resolved. 



While the Wahba problem is generally defined for vector observations of arbitrary dimension, including both unit and non-unit vectors, we focus on 3D unit vectors. In this case, the observations can be interpreted as points on the unit sphere S2. For example, in \cref{fig:onecol_overview}(b), top row, two point clouds of a guitar—source (blue) and target (red)—are shown on the left. In the middle, these 3D point clouds are encoded as 3D unit vectors from the origin using the Extended Gaussian Image (EGI) embedding \cite{EGIHorn}. We can estimate the rotation of this spherical embedding, then apply the resulting rotation to the original 3D point clouds to align them. Thus, estimating the rotation between two spherical patterns can be treated as a variant of the Wahba problem.

The problem of rotation estimation between two spherical patterns is usually approached using spherical cross-correlation of spectral coefficients \cite{makadia2003direct, makadia2004rotation, makadia2006rotation}. Typically, spherical patterns are transformed into the frequency domain using spherical harmonics. Spherical cross-correlation \cite{sorgi2004normalized} is then applied, where each potential rotation of the target set is evaluated against the reference set based on a correlation score that measures alignment. The rotation maximizing this cross-correlation score is identified as the optimal rotation. 

In general, spherical cross-correlation has shown promising results across various applications, including camera pose estimation \cite{makadia2006rotation}, point cloud registration \cite{makadia2006fully}, shape alignment \cite{gutman2008shape}, and localization using catadioptric cameras \cite{lin2021self}.  It has also been used in other areas such as analyzing 3D radiation patterns of musical instruments \cite{carpentier2023spherical} and spherical cortical surface registration \cite{zhao2020unsupervised,zhao2021s3reg}. Importantly, the development of spherical CNNs \cite{cohen2018spherical, esteves2023scaling} has relied heavily on spherical cross-correlation operations.

However, these methods have certain drawbacks. The accuracy of spherical cross-correlation methods is resolution-dependent on the rotation space \cite{makadia2006rotation}. The computational complexity of these methods is \(\mathcal{O}(N_r^3 \log(N_r))\), where \(N_r\) represents the sampling of the rotation space \cite{sorgi2004normalized}, constrained by the bandwidth, or the number of spherical harmonic coefficients retained \cite{makadia2006rotation}. This complexity restricts the scalability of methods that depend on spherical cross-correlation \cite{esteves2023scaling}. Furthermore, limited studies have examined performance under conditions of high rotation errors, where the entire \(SO(3)\) space may need to be explored. For instance, in \cite{makadia2006rotation}, the authors found their method performed well for source rotation errors (Euler angle \(\beta\)) around 60 degrees, but struggled beyond 75 degrees. 
Our work aims to address these limitations of spherical cross-correlation. Our contributions are outlined below:


\begin{itemize}
    \item We introduce two novel algorithms for spherical point-pattern registration, along with a third hybrid algorithm that combines the two.  These solve the Wahba problem in the presence of noise and outliers. Through extensive analysis on simulated data, we 
    demonstrate how the hybrid algorithm effectively addresses these limitations.
    
    \item We show that our algorithms operate with \(O(n)\) time complexity, ensuring both speed and scalability.
    
    \item We demonstrate the adaptability of our algorithms for point cloud registration, encompassing both complete-to-complete and partial-to-complete tasks. Additionally, we present the Centroid Aware Spherical Embedding (CASE) method, which converts 3D object point clouds onto a unit sphere to facilitate point cloud registration.
    
    \item We propose a novel approach for converting spherical images to spherical point clouds, enabling tasks such as rotation estimation between two spherical images. Our method can accurately estimate rotation even when one image undergoes large rotations and contains clutter.

    \item We present the ``Robust Vector Alignment Dataset,'' which includes 5 template spherical patterns and 700 source patterns per template (a total of 3500 patterns) generated by 100 rotations randomly sampled over the entire rotation space (\(SO(3)\)), with varying levels of noise and outliers.
    
\end{itemize}


%% file: sec/2_Related_work.tex
\section{Related Work}
\label{sec:related_work}

A vast literature addresses the Wahba problem. In the simplest case—where there is no noise, no outliers, and true one-to-one correspondences are provided—closed-form solutions exist for both the Wahba problem \cite{arun1987least, forbes2015linear, horn1987closed, horn1988closed, khoshelham2016closed, markley1988attitude, saunderson2015semidefinite} and the Orthogonal Procrustes Problem \cite{schonemann1966generalized}, where orthogonal matrices are sought instead of a rotation matrix.

When vector observations are noisy or when the source set contains more observations than the reference set, one common practice is to estimate correspondences using various feature descriptors \cite{rusu2009fast, tombari2013performance, drost2010model, choy2019fully}.  However, falsely-identified correspondences can introduce inaccuracies.

Local optimization methods are often used to solve these correspondence-based approaches \cite{besl1992method, chetverikov2002trimmed, antonante2021outlier, yang2019quaternion, chui2003new, jian2010robust, myronenko2010point, rusinkiewicz2001efficient, yang2020graduated, peng2022arcs}, as well as robust methods based on RANSAC \cite{fischler1981random, li2020gesac, li2021point, sun2021ransic}. However, the success of these methods largely depends on the quality of correspondences detected by front-end feature detectors \cite{Talak23}, and detecting robust correspondences is computationally costly for larger point clouds. To address this, some newer approaches focus on outlier removal \cite{parra2015guaranteed, parra2019practical, bustos2017guaranteed, shi2021robin}; however, they still rely heavily on the quality of correspondences. 

In contrast, global optimization approaches employ correspondence-free methods \cite{campbell2016gogma, chin2016guaranteed, li2011weak, lian2016efficient, liu2018efficient, parra2014fast, straub2017efficient, yang2015go}. The primary limitation of these approaches is their high runtime, making them infeasible for real-time applications.

In the domain of geometric rotation search, other methods, such as spherical cross-correlation, orientation histograms, and FFT-based approaches, have been explored \cite{makadia2006fully, makadia2004rotation, bernreiter2021phaser, ma2016fast}. Spherical correlation-based methods struggle 
in real-time applications primarily because existing spherical correlation techniques have higher computational complexity than planar cross-correlation methods \cite{esteves2023scaling, sorgi2004normalized}. 

Additionally, recent methods based on deep learning have emerged \cite{huang2021predator, choy2020deep, bai2021pointdsc, bauer2021reagent, yew2020rpm, sarode2019pcrnet, aoki2019pointnetlk}. While these neural network-based models---often designed to mimic the ICP algorithm or learn global representations---have shown improved performance, some models still struggle to generalize under conditions with occlusions and partial views \cite{Talak23}.

%% file: sec/3_Mathematical_preliminaries.tex
\section{Concepts and Preliminaries}
\label{sec:mathematical_Preliminaries}

In this section, we clarify the mathematical preliminaries needed to formally define our algorithms. The Supplementary Material (Sec.~\ref*{supplementary_sec:preliminaries}) provides background on Rotations as \( \text{SO(3)} \), Rotations between vectors, Axis Direction Angles, and the Rotation Matrix from Axis-Angles.

\textbf{S2-Points:} S2-points represent a set of 3D points \( x \! \in \! \mathbb{R}^3 \) located on the surface of a unit sphere, each with a norm of 1. In the spherical coordinate system, these points can be parameterized by the azimuthal angle \( \alpha \in [0, 2\pi] \) and the polar angle \( \beta \in [0, \pi] \), with a fixed radius of 1: 
%
\begin{equation}
\begin{aligned}
\begin{rcases}
\alpha_{\text{rad}} &= \text{atan2}(y, x), \quad \beta_{\text{rad}} = \text{acos}(z), \\
\alpha_{\text{deg}} &= \alpha_{\text{rad}} \left( \frac{180}{\pi} \right), \quad \beta_{\text{deg}} = \beta_{\text{rad}} \left( \frac{180}{\pi} \right)
\label{3D_to2D}
\end{rcases}
\end{aligned}
\end{equation}

\begin{figure}[t!]
  \centering
   \includegraphics[width=1\linewidth]{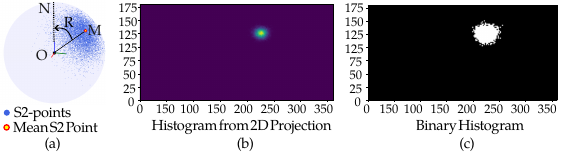}
   \caption{Spherical Point to Binary Histogram. (a) Spherical point with `Mean Direction' \(\overrightarrow{OM}\), `North Direction' \(\overrightarrow{ON}\), and \(R\) as the rotation matrix between the two vectors. (b) Histogram (Probability Density Map) created by projecting S2 points into 2D. (c)~Resulting Binary Histogram.}
   \label{fig:histogram}
\end{figure}

\textbf{2D Histogram Representation of S2-Points: }
\label{binary_histogram}
We use the inherent structure of the spherical coordinate system for S2-points to formulate a two-dimensional histogram representation. Specifically, each S2-point is projected onto an equilateral projection plane, from which a 2D histogram is constructed. For noisy data, a coarser bin resolution yields slightly better performance, whereas for dense point clouds, a finer resolution is preferable. As such, we adopt a bin resolution of \(360 \times 180\), corresponding to \(1^\circ\) increments in both azimuth and elevation angles. This configuration optimally balances speed and accuracy across most of our datasets. In its basic form, this histogram functions as a probability density map, as shown in \cref{fig:histogram}(b). To focus on overall distribution rather than isolated high peaks, we use a parameterized binary histogram representation, shown in \cref{fig:histogram}(c). Here, we apply a threshold: bins containing a count of points greater than or equal to this threshold are assigned a value of 1, while bins below this threshold are assigned 0. This parameterization allows us to capture the overall spatial pattern of points without an overemphasis on regions with high point density. Since our point clouds had many empty bins, we used a threshold of 0 for those.

\textbf{Mean Direction and North Direction:} 
\label{mean_direction}
The ``mean direction" of S2-points refers to the central point on the unit sphere that best represents the average direction of all data points, 
effectively acting as a centroid when considering directional data \cite{mardia2009directional,ley2017modern}. This centroid generally does not lie on the surface of the sphere. However, we are interested in the mean direction vector, which is the unit vector passing through the centroid of the S2-points. In \cref{fig:histogram}(a), the vector \( \overrightarrow{OM} \) represents the mean direction vector of the S2-points, and the point \( M \), which lies on the sphere, is simply referred to as the ``mean S2-point."

The ``North direction vector'' is the direction of the absolute north, usually towards the $z$-azis of the globe (more specifically where where latitude angle is $+90^\circ$). In \cref{fig:histogram}(a), \( \overrightarrow{ON} \) is the ``North direction vector.''

%% file: sec/4_Methods.tex
\section{Spherical Point Pattern Registration}
\label{sec:Methods}

The problem of spherical point pattern registration can be formulated as follows: given two S2-point sets, \( \mathcal{A} = \{\mathbf{a}_i\}_{i=1}^{N} \) and \( \mathcal{B} = \{\mathbf{b}_j\}_{j=1}^{M} \) (where \( \{\mathbf{a}_i\}, \{\mathbf{b}_j\} \in \mathbb{S}^2 \subset \mathbb{R}^3 \)), how can we find the rotation matrix \( \mathbf{R} \) that best aligns the source set \( \mathcal{B} \) to the template set \( \mathcal{A} \)?

In the simplest case, let us assume \( M = N \), meaning sets \( \mathcal{A} \) and \( \mathcal{B} \) contain the same number of points, and that there exists a one-to-one correspondence between points in \( \mathcal{A} \) and \( \mathcal{B} \). In this case, we have \( \mathbf{b}_i = \mathbf{R} \mathbf{a}_i \), where \( \mathbf{R} \) is the desired rotation matrix. This problem can be efficiently solved using closed-form solutions, as 
in \cite{kabsch1976solution, horn1987closed, arun1987least}.

However, we consider the problem under more challenging conditions, where no explicit correspondence is assumed, and the data contain random outliers (i.e., \( M \neq N \)). This formulation is similar to the ``Robust Wahba Problem,'' as described in \cite{yang2019quaternion}, with the added complexity that point correspondences are unknown. 

In this section, we introduce three algorithms: two novel algorithms designed to solve the spherical point pattern registration problem, and a third that combines the first two. 

\subsection{Algorithm 1: Spherical Pattern Matching by Correlation (SPMC)}
\label{algo1}
    \textbf{Step 1:} Compute the rotation matrices \( R_A \) and \( R_B \) that align the mean direction of each set, \(\mathcal{A} = \{\mathbf{a}_i\}_{i=1}^{N}\) and \(\mathcal{B} = \{\mathbf{b}_j\}_{j=1}^{M}\), with the North Pole direction \([0, 0, 1]\).

    \vspace{2pt}
    \noindent
    \textbf{Step 2:} Using the rotation matrices \( R_A \) and \( R_B \), rotate each point in \(\mathcal{A}\) and \(\mathcal{B}\) to align the sets with the North Pole. For each point \(\mathbf{a}_i \in \mathcal{A}\) and each point \(\mathbf{b}_j \in \mathcal{B}\), we define:
    \[
    \mathcal{A}^{\text{NP}} = \{ R_A \, \mathbf{a}_i \}_{i=1}^{N}, \quad \mathcal{B}^{\text{NP}} = \{ R_B \, \mathbf{b}_j \}_{j=1}^{M}
    \]
    where \(\mathcal{A}^{\text{NP}}\) and \(\mathcal{B}^{\text{NP}}\) represent the transformed points in \(\mathcal{A}\) and \(\mathcal{B}\) after rotation to align with the North Pole.
    
    Optionally, we can normalize \(\mathcal{A}^{\text{NP}}\) and \(\mathcal{B}^{\text{NP}}\), aligning each vector toward the positive \(z\)-axis. This step is particularly helpful when dealing with surface normals, as it minimizes orientation discrepancies by ensuring that all vectors are directed consistently.

    \vspace{2pt}
    \noindent
    \textbf{Step 3:} Compute the binary histograms \(\mathcal{H}_A\) and \(\mathcal{H}_B\) from the aligned sets \(\mathcal{A}^{\text{NP}}\) and \(\mathcal{B}^{\text{NP}}\), respectively. Here, \(\mathcal{H}_A\) serves as the template histogram and \(\mathcal{H}_B\) as the source histogram (see \cref{binary_histogram} for details).

   \vspace{2pt}
    \noindent
    \textbf{Step 4:} Perform a 1D circular cross-correlation between the fixed histogram \(\mathcal{H}_A\) and the moving histogram \(\mathcal{H}_B\). As discussed in \cref{binary_histogram}, we use histograms with 360 horizontal bins (representing the azimuthal angular range from \(0^\circ\) to \(360^\circ\)) and 180 vertical bins (representing the polar angular range from \(0^\circ\) to \(180^\circ\)).
   Convert the 2D histograms \(\mathcal{H}_A\) and \(\mathcal{H}_B\) into 1D arrays \( h_A \) and \( h_B \) by summing over the vertical bins (polar angle), resulting in 1D representations with 360 bins for each azimuthal angle. Define the 1D circular cross-correlation \( C(s) \) as:
   \begin{equation}
   C(s) = \sum_{\lambda=0}^{359} h_A(\lambda) \cdot h_B \bigl( (\lambda + s) \!\! \mod 360 \bigr)
   \end{equation}
   where \( s \) is the circular shift applied to \( h_B \) along the azimuthal angle, ranging from \(0\) to \(359\).
   Then, find the optimal shift \( s^* \) that maximizes the circular cross-correlation:
    \begin{equation}
    \label{eqn:shift_opt}
    s^* = \arg \max_{s} C(s)
    \end{equation}
    
   \noindent
   \textbf{Step 5:} Compute the desired rotation using:

    \begin{equation}
    R_{\text{shift}} = R_z(s^*)
    \end{equation}
    \noindent
    where \( R_z(\theta) \) is the rotation matrix about the \(z\)-axis by angle \(\theta\) as shown in \cref{eqn:rxyz}.
    Then, the optimal rotation \( R_{\text{opt}} \) that aligns \(\mathcal{B}\) with \(\mathcal{A}\) is given by:

    \begin{equation}
    \label{final_rotation_algo_1}
        R_{\text{opt}} = R_A^{-1} \cdot R_{\text{shift}} \cdot R_B
    \end{equation}

\begin{figure}[t]
  \centering
   \includegraphics[width=1\linewidth]{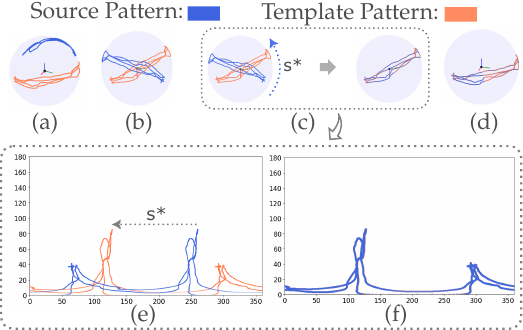}
   \caption{Overview of Spherical Pattern Matching by Correlation (SPMC). (a) Source and template patterns, with the source pattern initially rotated. (b) Both source and template patterns are aligned to the north pole of the sphere (Step 2). (c) With the means aligned, the source pattern is rotated around the azimuthal axis relative to the north vector until it overlaps with the template. (d) Performing 1D cross-correlation: rotating around the azimuthal axis corresponds to a 1D cross-correlation along the azimuthal direction of the binary histogram. (f) The patterns are aligned at the position where maximum correlation is achieved.}
   \label{fig:onecol}
\end{figure}

An overview of the algorithm is shown in \cref{fig:onecol}, with an example of a one-to-one (closed-form solution) case. A formal proof of this case is provided in the Supplementary Material, Sec.~\ref*{supplementary_sec:proof}.

\subsection{Algorithm 2: Fast Rotation Search (FRS)}

At a high level, the Fast Rotation Search (FRS) algorithm is an iterative optimization-based method designed to compute the required rotation matrix. The detailed algorithm is provided in \cref{alg:pseudocode}. The process begins by calculating fixed histograms from the axis direction angles of the target set \(\mathcal{A}\). At each iteration, moving histograms are then computed from the axis direction angles of the source set \(\mathcal{B}\). The algorithm performs a 1D circular cross-correlation between the fixed histograms and the corresponding moving histograms to determine the optimal shift, as described in \cref{eqn:shift_opt}. From this optimal shift, an intermediate rotation matrix is calculated. The source is rotated by this intermediate rotation, and the process is repeated, generating updated moving histograms and recalculating the intermediate rotation matrix until the moving and fixed histograms optimally align. Despite being an iterative method, the FRS algorithm operates in near-linear time; its computational complexity is discussed in \cref{sec:computationalcomplexity}.


\setcounter{algorithm}{1}
\begin{algorithm}[ht]
\caption{FRS: Fast Rotation Search}
\label{alg:pseudocode} 
\textbf{Input:} Target spherical point cloud $\mathcal{A}$, source spherical point cloud $\mathcal{B}$, bin multiplier $k=1$, max iterations $= 50$, target shift $s_T = 0$ \\
\textbf{Output:} Rotation matrix $R_{\text{est}}$ aligning $\mathcal{B}$ to $\mathcal{A}$
\begin{algorithmic}[1]
    \State Compute \textbf{axis direction angles} of $\mathcal{A}$ \Comment{$N \times 3$ array of angles}
    \State Compute fixed 1D histograms for $\mathcal{A}$ with $k \times 360$ bins. Generate three fixed histograms, one for each axis direction angle.
    \State Initialize $\texttt{Set\_B} \gets \mathcal{B}$, $i \gets 0$
    
    \While{$i < \texttt{max\_iterations}$}
        \State Compute \textbf{axis direction angles} of $\texttt{Set\_B}$
        \State Compute 1D histograms for $\texttt{Set\_B}$ along each axis
        \State Perform 1D circular cross-correlation between the fixed and moving histograms to find $\texttt{x\_shift}$, $\texttt{y\_shift}$, $\texttt{z\_shift}$
        \State Compute rotation matrices $R_{x\text{shift}}$, $R_{y\text{shift}}$, $R_{z\text{shift}}$ based on shifts
        \State Combine rotations $R \gets R_z \cdot R_y \cdot R_x$
        \If{$\texttt{x\_shift} = s_T$ and $\texttt{y\_shift} = s_T$ and $\texttt{z\_shift} = s_T$}
            \State \textbf{break}
        \EndIf
        \State Update $\texttt{Set\_B} \gets \texttt{rotate}(\mathcal{B}, R)$
        \State $i \gets i + 1$
    \EndWhile
    \State Compute $R_{\text{est}}$ using closed-form solution from initial and final $\texttt{Set\_B}$
    \State \textbf{return} $R_{\text{est}}$
\end{algorithmic}
\end{algorithm}

\subsection{Algorithm 3: SPMC+FRS}

We also propose a hybrid algorithm that combines Algorithm 1 and Algorithm 2. This approach is straightforward: we first perform the SPMC algorithm, which serves as an initialization for the FRS algorithm, and then proceed with the FRS algorithm to complete the alignment.

\subsection{Computational Complexity}
\label{sec:computationalcomplexity}
The following is a breakdown of the computational complexity of our algorithms.

\textbf{SPMC:} 
Referring to the algorithm described in \cref{algo1}, from Step 1 to Step 3: obtaining the mean directions, projecting to the North Pole, performing 2D projection, and histogram binning each have a computational complexity of \(O(n)\). Step 4, which involves performing a 1D cross-correlation, operates in constant time \(O(1)\), as the correlation is between histograms with a fixed number of bins. Additionally, the rotation estimation in \cref{final_rotation_algo_1} also runs in \(O(1)\). Therefore, the overall complexity of the SPMC algorithm is approximately \(O(n)\).

\textbf{FRS:} 
Computing the axis direction angles and histogram binning both operate in \(O(n)\) time. Subsequently, the algorithm employs a 1D cross-correlation procedure analogous to SPMC, iterating up to a maximum of \(K\) times (we set \(K=50\)). Although our experiments indicate that the algorithm converges on average in 11 iterations—with a maximum observed of 34 iterations—we conservatively choose \(K=50\) as an upper bound. More details are provided in the Supplementary Material, Sec.~\ref*{convergence-analysis}. Consequently, this step has a complexity of at most \(K \times O(1)\). Finally, estimating the final rotation via a closed-form solution also runs in \(O(n)\) time. Overall, the FRS algorithm operates in linear time, \(O(n)\).

\textbf{SPMC+FRS:}
Since both SPMC and FRS run at \(O(n)\), we can conclude that SPMC+FRS also runs at \(O(n)\).


%% file: sec/5_Experiments.tex
\section{Experiments}
\label{sec:Experiments}

\subsection{Experiment 1: Robust Alignment}
\label{simulation_exp}

Sec.~\ref*{supplementary-exp-1} in the Supplementary Material provides an explanation of the datasets and methods for this experiment.  Briefly, datasets \(A_1\)-\(A_5\) are five spherical patterns with different features such as islands, lines, and non-uniform densities; noise patterns \(B_1\)-\(B_7\) are added, then these are rotated in 100 different random orientations.  We compare our algorithms against FPFH+QUASAR, where correspondences are provided by FPFH \cite{rusu2009fast} and QUASAR \cite{yang2019quaternion} conducts registration.  QUASAR is used because it achieved state-of-the-art results for the Wahba problem, even with up to 95\% outliers, when correspondences were given.

\begin{figure*}[ht!]
  \centering
   \includegraphics[width=1.0\linewidth]{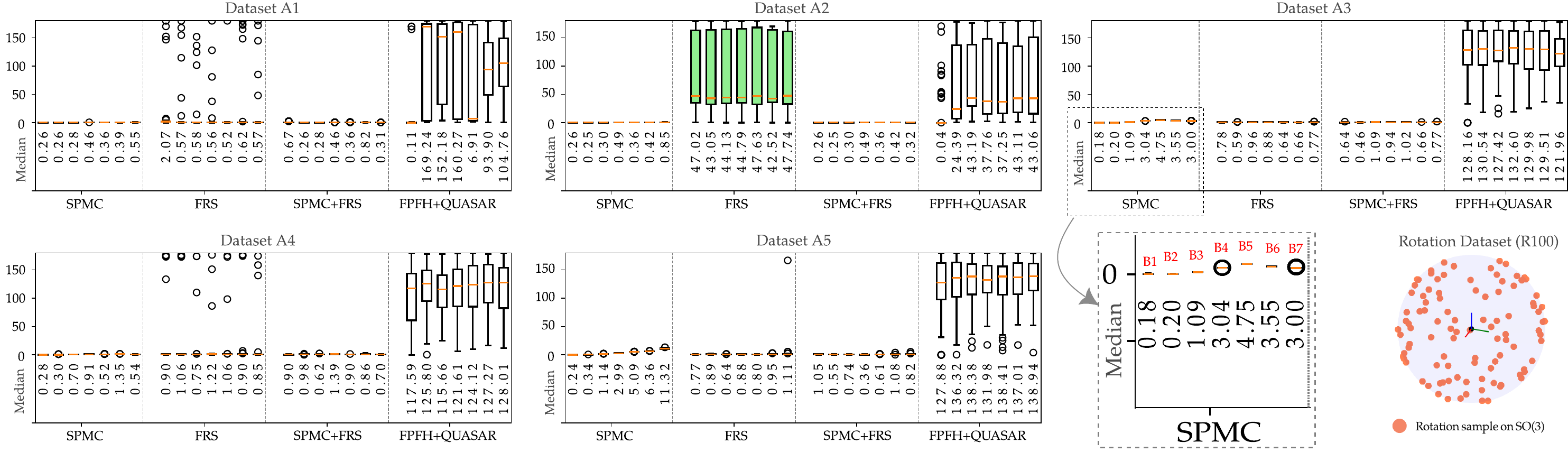}
   \caption{Quantitative results of our algorithms and FPFH+QUASAR on the ``Robust Vector Alignment Dataset'' and Rotation Dataset (R100). For each dataset, both methods display all 7 source set configurations in box plots. The median value of each box plot is noted below each plot. The 7 configurations, labeled \(B1\) through \(B7\), are arranged from left to right, as shown in the zoomed-out section of the plot for Dataset 3. In the bottom right, the distribution of 100 rotations (R100) over the \(SO(3)\) space is displayed.}
   \label{fig:sim_quant}
\end{figure*}

\begin{figure}[ht!]
  \centering
   \includegraphics[width=1\linewidth]{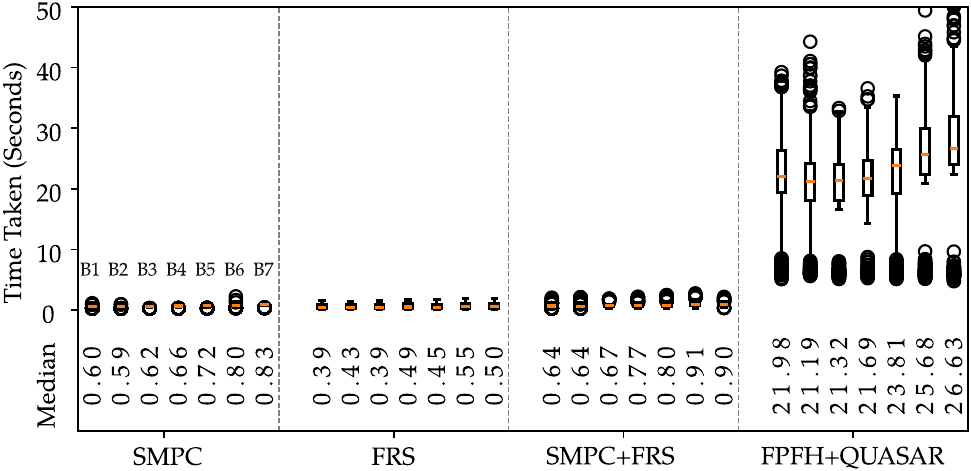}
   \caption{Analysis of time taken of our algorithms and FPFH+QUASAR on ``Robust Vector Alignment Dataset". The median value of each box plot is written below each plot. For each method results of the 7 configurations, labeled \(B1\) through \(B7\), are arranged from left to right, as shown in the result of SMPC.}
   \label{fig:sim_time}
\end{figure}

\textbf{Results and Analysis of Robust Alignment:}  
As shown in \cref{fig:sim_quant}, SPMC+FRS consistently achieves the best performance across all datasets, with a median rotational angle error of less than \(1^\circ\) when combining all datasets. In contrast, the FPFH+QUASAR results reveal challenges in finding the optimal rotation: while the backend can be certifiably optimal, they may still fail if the  correspondence detection produces inaccurate results \cite{Talak23}. 

The FPFH+QUASAR results also underscore the dataset-dependent difficulty of finding correspondences in spherical data. For example, in the one-to-one cases for datasets A1B1 and A2B1, FPFH+QUASAR achieves a median angular error of $0.11^\circ$ and $0.04^\circ$, respectively. However, for other one-to-one cases like A3B1, A4B1, and A5B1, median errors exceed $100^\circ$. This discrepancy reflects the challenges feature detectors face with spherical point clouds: unlike object point clouds with multidimensional depth and non-uniform surfaces, spherical point clouds on a unit sphere have a uniform surface and lack depth, complicating correspondence detection. In contrast, SPMC+FRS consistently finds near-optimal rotations across all datasets, demonstrating its robustness.

In Dataset \(A_2\), FRS alone performs sub-optimally, whereas SPMC+FRS achieves excellent registration across all rotation combinations. This suggests that the FRS algorithm alone is sensitive to initialization, especially for shapes with sharp features and localized points (e.g. concentrated on only one side of the sphere).

In Dataset \(A_5\), the accuracy of SPMC decreases (median error rising from \(0.24^\circ\) to \(11.32^\circ\)) as the number of outliers increases (from \(B_1\) to \(B_7\)). This is due to SPMC aligning the mean of the source and destination clouds at the North Pole. When there are many (e.g., 90\%) added outliers, the mean shifts toward the outliers, affecting the alignment.

As shown in \cref{fig:sim_time}, our algorithms operate in under 1 second across all cases, whereas FPFH+QUASAR has a median computation time of approximately 23.18 seconds. This suggests that our algorithm can efficiently handle large point clouds without compromising accuracy.

\begin{figure}[ht]
  \centering
   \includegraphics[width=0.85\linewidth]{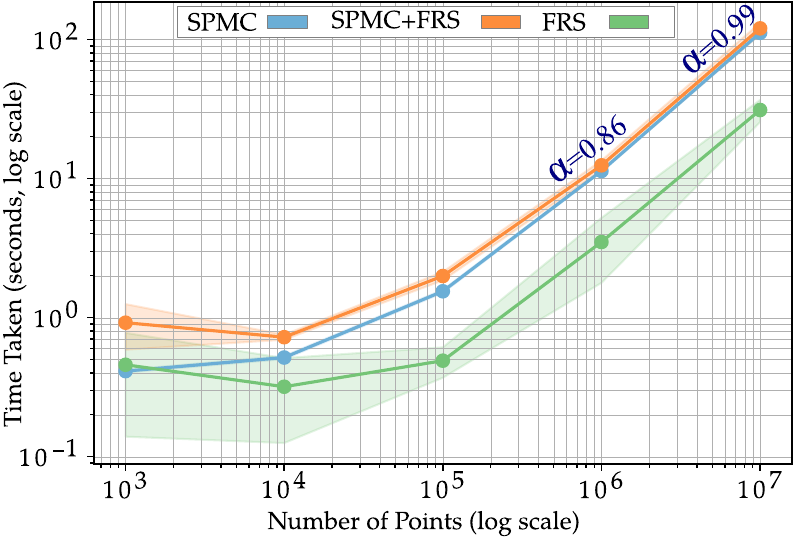}
   \caption{Computational Complexity of our algorithms. Time taken versus the number of points in the source set (B3) and template set (A3), evaluated across 20 rotations. The X-axis represents the number of source and template points, while the Y-axis displays the time taken in logarithmic scale. $\alpha$ is the empirical slope of runtime scaling for SPMC.}
   \label{fig:computation_complexity}
\end{figure}

\textbf{Experimental Computational Complexity:}
In \cref{fig:computation_complexity}, we experimentally show the computational complexity of our method. For the template cloud, we use Dataset \(A_3\) and resample points at different scales ranging from \(10^3\), \(10^4\), \(10^5\), \(10^6\), to \(10^7\). We then generate source sets by applying 20 random rotations to each scale. The results show that for larger point sets (\(10^5\), \(10^6\), and \(10^7\)), the plot exhibits near-linear time complexity \(O(n)\). There is some variability in FRS due to fluctuations in the number of iterations required. 

\subsection{Experiment 2: Point Cloud Registration}

In this experiment, we implemented our algorithm for both complete-to-complete (Comp2Comp) and partial-to-complete (Part2Comp) point cloud registration with the Modelnet40 dataset \cite{wu20153d}. Additional details and comparisons with the 3DMatch \cite{zeng20173dmatch} and KITTI \cite{geiger2013vision} datasets are in the Supplementary Material, Sec.~\ref*{supplementary-exp-2}.

\textbf{Spherical Embedding for Point Cloud:}  
%
Our method starts with representing the point cloud in a robust spherical embedding, such as the Extended Gaussian Image (EGI) \cite{EGIHorn}, which maps surface normals onto the sphere. Normals are computed by fitting planes to local neighborhoods found via a KD-tree (with a specified search radius), and then refined using to ensure normal consistency across the surface. We use EGI in our method and additionally use the ``Centroid Aware Spherical Embedding" (CASE), which presumes a known object centroid. CASE connects lines from the centroid to each point, projecting them onto the unit sphere. This embedding, invariant to scale, rotation, translation, and capable of handling more than 50\% outliers and 0.01 Gaussian noise, can show benefits relative to EGI.
(CASE is further discussed in Supplementary Material Sec.~\ref*{supplementary-CASE}). We perform Comp2Comp registration using both CASE and EGI, and partial-to-complete registration using only EGI.

The rationale for CASE is that, in the Comp2Comp problem, the centroid remains known as both the source and target shapes are complete, removing the need for centroid estimation. However, for partial-to-complete registration, estimating the centroid using only a partial or occluded view is challenging. Some recent studies have shown promising results in estimating the centroid of objects with partial point clouds \cite{zhao2021centroidreg}, but we have not explored this approach. 

\textbf{Translation Estimation:}  
In Comp2Comp registration, translation can be estimated by aligning the centroids of the source and target point clouds, assuming no change in centroids. To enhance robustness and account for uncertainties in centroid estimation in Part2Comp, we employ a voxel-based adaptive voting scheme for coarse translation estimation, followed by a translation-only ICP for fine alignment. Details are in the Supplementary Material in Sec.~\ref*{supplementary-translation-estimation}. 
 Unlike point-to-point adaptive voting as in \cite{yang2020teaser}, our voxel-based approach aggregates point information within grid cells, improving both robustness and computational efficiency.
This algorithm assumes that the source point cloud has been pre-aligned in rotation with the target. In other words, our approach decouples rotation and translation estimation: it performs rotation first, then translation.

\begin{figure*}[ht!]
  \centering
   \includegraphics[width=1\linewidth]{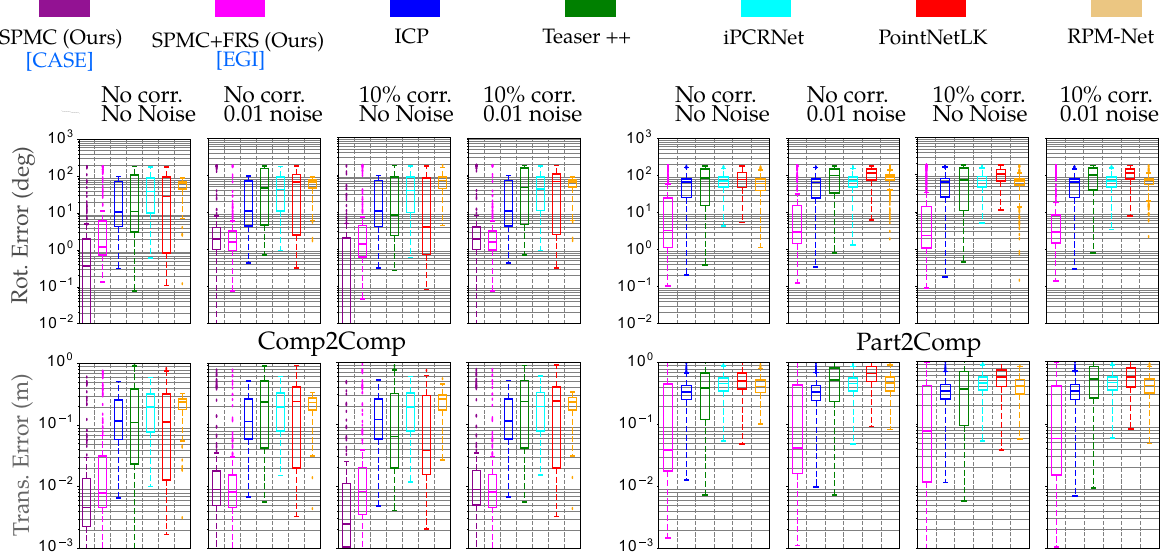}
   \caption{Modelnet40 quantitative results: Rotational error in degrees (top row) and translation error in meters (bottom row) for both complete to complete (Comp2Comp) and partial to complete (Part2Comp) in 4 cases with different amounts of noise and correspondence.}
   \label{fig:pcr_quant}
\end{figure*}

\textbf{Results of Point Cloud Registration:}
\cref{fig:pcr_quant} presents a detailed evaluation of point cloud registration for both the Comp2Comp and Part2Comp datasets. We compared our algorithm with ICP, TEASER++ (using FPFH for correspondence), and deep learning models, including PCRNet, PointNetLK, and RMN-Net. For the deep learning models, we used pretrained weights from the ModelNet40 dataset \cite{learning3d}, where each model was sampled with 1024 points.

In Comp2Comp, SPMC with CASE spherical embedding achieved the best results in both rotation and translation error, with a median rotational error of \(0.13^\circ\). In contrast, TEASER++ (using FPFH for correspondence) and ICP had median errors around \(10^\circ\), indicating their dependence on initialization and the quality of correspondences. In the ``No Correspondence" case, the median rotational error of TEASER++ increased when noise was introduced. Notably, there was little difference in TEASER++ performance between the ``No Corr., No Noise" case and the ``10\% Corr., No Noise" case, highlighting the challenges of reliably identifying correspondences even with a small percentage of true correspondences. TEASER++ performance decreased significantly when noise was added, underscoring the increased difficulty of identifying correspondences under noisy conditions. The deep learning models also performed poorly, revealing their limitations in generalizing to configurations with different numbers of input points (2500 points) compared to their training points (1024 points).

In the challenging Part2Comp case, SPMC+FRS achieved rotational errors around \(2^\circ\), while other methods struggled to reach good registration. Both rotational and translation errors were consistent for our method, showing the robustness of our algorithm for pose estimation, even in the presence of occlusion.

\subsection{Experiment 3: Rotation estimation from spherical images}

\label{experiment3}

In this experiment, we demonstrate how our algorithm can be adapted for rotation estimation from spherical images, and related tasks \cite{makadia2003direct, makadia2004rotation, makadia2006rotation}. 


\textbf{Dataset:} For this experiment, we used a simple 2D world map image projected onto a unit sphere, which served as the template image. For convenience, the 2D image was resized to a pixel size of \(180 \times 360\). We then created five source sets (\( \text{Img1}, \ldots, \text{Img5} \)) by adding clutter (random shapes with different colors) at varying percentages, ranging from 0\% to approximately 19.3\%, across the entire sphere. This addition of clutter simulates, for instance, the presence of new objects in the scene as the camera rotates. The cluttered source sets were then rotated by 100 random rotations from the R100 set, as shown in \cref{fig:sim_quant}. A sample of the dataset and qualitative results are provided in the Supplementary Material, Sec.~\ref*{supplementary-exp-3}.


\textbf{Spherical Image to Spherical Points (SphImg2SphPoints):} For each pixel in the spherical image, we compute its intensity, ranging from 0 to 1. To binarize these values, we set a threshold intensity level: pixels with intensity below the threshold are assigned a value of 0, while those above are assigned a value of 1. A point is then assigned to each pixel with an intensity of 1. A lower threshold increases the number of points, potentially recovering more overlapping features but also capturing more noise and clutter. Conversely, a higher threshold yields fewer points, which can challenge registration if insufficient features are captured. For this experiment, we set an intensity threshold of 0.21. Details about the intensity threshold are in the Supplementary Material.

\begin{figure}[t!]
  \centering
   \includegraphics[width=0.97\linewidth]{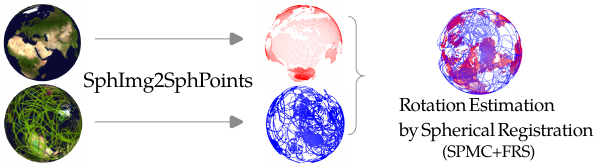}
   \caption{Rotation estimation from spherical images. First, the template spherical image (top left) and source spherical image (bottom left) are converted from image space to spherical point clouds using SphImg2SphPoints. Then, our algorithm (SPMC+FRS) is applied to perform registration between the two sets of spherical points.}
   \label{fig:sph_method}
\end{figure}

Once the spherical image is converted to a spherical point cloud, we apply the SPMC+FRS algorithm for registration and rotation estimation (\cref{fig:sph_method}).

\begin{figure}[t!]
  \centering
   \includegraphics[width=0.7\linewidth]{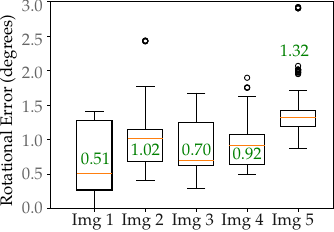}
   \caption{Results of rotation estimation from spherical images, showing the angular errors. Each source spherical image (Img1 to Img5) was initialized with 100 rotations over \(SO(3)\), and the results represent the error after rotation estimation using our algorithm. Median errors are highlighted in green.}
   \label{fig:sph_quant_res}
\end{figure}

\textbf{Results of Rotation Estimation From Spherical Images:} \cref{fig:sph_quant_res} presents the quantitative results of rotation estimation from the five spherical image sets under 100 random rotations. Across all sets, we achieved a median angular error of \(0.89^\circ\). Notably, for Img5 (with 19.3\% clutter), we observed a median error of \(1.32^\circ\) and a peak error of \(2.96^\circ\). Our algorithm is thus robust.

%% file: sec/6_Conclussion.tex
\section{Conclusion}
\label{sec:Conclusion}


We have presented novel methods for the rotation estimation of spherical patterns. Our algorithms are both fast and robust, operating in linear time. Extensive simulations demonstrate that our approach successfully aligns patterns even under high outlier ratios. To our knowledge, no prior work has achieved spherical cross-correlation with this combination of speed and accuracy. Furthermore, we show how our algorithms can be adapted for applications in point cloud registration and spherical image registration.



The current work focuses exclusively on the 3D case; however, the proposed method is readily extensible to unit vector alignment in higher dimensions. Also, We do not explicitly address the design of robust spherical embeddings for point clouds, which remains a critical factor in registration performance. While EGI captures object pose effectively and may benefit learning-based tasks such as rotation representation, developing more robust spherical embeddings and extending the framework to 
N-dimensional settings is left for future work.

\noindent
\textbf{Acknowledgement:} This work was funded by the National Science Foundation, Grant \# 2014499.



%% file: sec/X_suppl.tex
\clearpage
\setcounter{page}{1}
\maketitlesupplementary

\section{Mathematical Preliminaries}
\label{supplementary_sec:preliminaries}

This section contains additional mathematical background needed for the algorithms in our paper.

\subsection{Rotations}  
The set of all rotations in three dimensions is denoted as \( \text{SO(3)} \), known as the ``special orthogonal group." Rotations can be represented by \( 3 \times 3 \) matrices that preserve both distance (i.e., \( \| \mathbf{R}\mathbf{x} \| = \| \mathbf{x} \| \)) and orientation (i.e., \( \det(\mathbf{R}) = +1 \)). If we represent points on the sphere as 3D unit vectors \( \mathbf{x} \), a rotation can be applied using the matrix-vector product \( \mathbf{R}\mathbf{x} \) \cite{cohen2018spherical}. 

\subsection{Rotation Between Two Vectors}  
\label{rot_vec}
To align one unit vector \( \mathbf{v_2} \) with another unit vector \( \mathbf{v_1} \), we can use the Rodrigues rotation matrix formula:

\begin{equation}
\mathbf{R} = 
\begin{cases}
\mathbf{I}, & \text{if } \mathbf{\hat{v}_1} = \mathbf{\hat{v}_2} \\
-\mathbf{I} + 2 \mathbf{v} \mathbf{v}^T, & \text{if } \mathbf{\hat{v}_1} = -\mathbf{\hat{v}_2} \\
\mathbf{I} + \mathbf{V} \sin \theta + \mathbf{V}^2 (1 - \cos \theta), & \text{otherwise}
\end{cases}
\end{equation}

where:
\[
\mathbf{\hat{v}_1} = \frac{\mathbf{v_1}}{\|\mathbf{v_1}\|}, \quad \mathbf{\hat{v}_2} = \frac{\mathbf{v_2}}{\|\mathbf{v_2}\|} ,
\mathbf{v} = \frac{\mathbf{\hat{v}_2} \times \mathbf{\hat{v}_1}}{\|\mathbf{\hat{v}_2} \times \mathbf{\hat{v}_1}\|}.
\]
\[
\theta = \arccos(\mathbf{\hat{v}_2} \cdot \mathbf{\hat{v}_1}),
\mathbf{V} = \begin{bmatrix} 0 & -v_z & v_y \\ v_z & 0 & -v_x \\ -v_y & v_x & 0 \end{bmatrix}.
\]

\subsection{Axis Direction Angles}
\label{axis_dir}

\begin{figure}[ht!]
  \centering
   \includegraphics[width=1\linewidth]{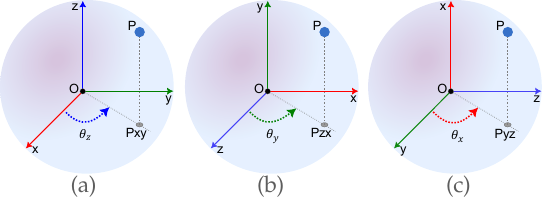}
   \caption{Spherical Point (P) to Axis Direction Angles. Projections of point \(P\) onto the \(xy\), \(zx\), and \(yz\) planes are denoted as \(P_{xy}\), \(P_{zx}\), and \(P_{yz}\), respectively. (a) \(\theta_z\) is the angle around the \(z\)-axis from the \(x\)-direction to the projection vector \(\overrightarrow{OP_{xy}}\). (b) \(\theta_y\) is the angle around the \(y\)-axis from the \(z\)-direction to the projection vector \(\overrightarrow{OP_{zx}}\). (c) \(\theta_x\) is the angle around the \(x\)-axis from the \(y\)-direction to the projection vector \(\overrightarrow{OP_{yz}}\).}
   \label{fig:axis_direction_angles}
\end{figure}

The axis direction angle of an S2-point 
around an axis is defined as the angle between the point’s projection onto the plane perpendicular to that axis and a reference axis. For example, in \cref{fig:axis_direction_angles}, the S2-point \( P \) is projected onto the perpendicular planes, and the axis direction angles are shown accordingly. Mathematically, these angles can be computed using the \( \textit{atan2()} \) function, as shown in \cref{eqn:axis_dir}. We set the range of the angle to lie between \( 0^\circ \) and \( 360^\circ \).

\begin{equation}
\label{eqn:axis_dir}
\begin{rcases}
    \theta_z &= \text{atan2}(y, x) \\
    \theta_y &= \text{atan2}(x, z) \\
    \theta_x &= \text{atan2}(z, y)
\end{rcases}
\end{equation}

\subsection{Rotation Matrix from Axis-Angles}

Given rotation angles \(\theta_x\), \(\theta_y\), and \(\theta_z\) around the \(X\)-, \(Y\)-, and \(Z\)-axes, respectively, we can compute the individual rotation matrices as follows:

\begin{equation}
\label{eqn:rxyz}
\setlength{\arraycolsep}{3pt} 
\begin{aligned}
R_z(\theta_z) &= \begin{bmatrix}
c_{\theta_z} & -s_{\theta_z} & 0 \\
s_{\theta_z} & c_{\theta_z} & 0 \\
0 & 0 & 1 
\end{bmatrix}, \quad
R_y(\theta_y) = \begin{bmatrix}
c_{\theta_y} & 0 & s_{\theta_y} \\
0 & 1 & 0 \\
-s_{\theta_y} & 0 & c_{\theta_y}
\end{bmatrix}, \\
R_x(\theta_x) &= \begin{bmatrix}
1 & 0 & 0 \\
0 & c_{\theta_x} & -s_{\theta_x} \\
0 & s_{\theta_x} & c_{\theta_x}
\end{bmatrix}, \;\; \text{[here } c\!: \cos(), s\!: \sin()\text{]}
\end{aligned}
\end{equation}

The combined rotation matrix \(R\) is then obtained by sequentially applying the rotations around each axis in the ZYX order:

\begin{equation}
R = R_z(\theta_z) \cdot R_y(\theta_y) \cdot R_x(\theta_x)
\end{equation}


\section{Formal Justification of SPMC:}  
\label{supplementary_sec:proof}
Centroid-based alignment is a classical strategy for estimating temporal shifts in noisy 1D signals, with theoretical performance bounds established in \cite{laguna1994time}. Our method is analogous to this idea in the spherical domain, where the azimuthal shift—after mean alignment and projection—corresponds to a 1D phase shift. In the absence of noise and outliers, aligning the mean directions of two spherical point sets ensures that the residual difference lies entirely in the azimuthal plane. A formal proof under this condition is below.

\begin{tcolorbox}[colback=white!97!gray, colframe=black!60!black,
title=Exact Recovery of SPMC (no noise or outliers), fonttitle=\bfseries, sharp corners=all, boxrule=0.4pt]

\textbf{Statement.} Let $\mathcal{B} = R_{\text{true}} \mathcal{A}$, where $\mathcal{A}, \mathcal{B} \in \mathbb{S}^2$ are identical spherical point clouds and $R_{\text{true}} \in \mathrm{SO}(3)$. Assume: (i) no noise or outliers, Then SPMC exactly recovers:
\[
R_{\text{opt}} = R_{\text{true}}. \vspace{-4pt}
\]

\textbf{Proof.} Let $\bar{\mathbf{a}}$ and $\bar{\mathbf{b}} = R_{\text{true}} \bar{\mathbf{a}}$ be the mean directions of $\mathcal{A}$ and $\mathcal{B}$. Let $R_A, R_B \in \mathrm{SO}(3)$ be rotations such that $R_A \bar{\mathbf{a}} = R_B \bar{\mathbf{b}} = [0,0,1]^T$. Then: \vspace{-8pt}
\[ 
\mathcal{A}^{\text{NP}} = R_A \mathcal{A}, \quad \mathcal{B}^{\text{NP}} = R_B R_{\text{true}} \mathcal{A} = R_{\text{res}} \mathcal{A}^{\text{NP}}, \vspace{-6pt}
\]
where $R_{\text{res}} = R_B R_{\text{true}} R_A^{-1}$ fixes the north pole and is therefore a $z$-axis rotation: $R_{\text{res}} = R_z(\theta)$. Azimuthal histogram correlation over $\mathcal{A}^{\text{NP}}$ and $\mathcal{B}^{\text{NP}}$ recovers $s^* = \theta$. The final estimate is: \vspace{-6pt}
\[
R_{\text{opt}} = R_A^{-1} R_z(s^*) R_B = R_{\text{true}}. \quad \blacksquare
\]
\end{tcolorbox}


\section{Experiment 1: Robust Alignment}
\subsection{Dataset Description}
\label{supplementary-exp-1}

In the first experiment, we evaluate the robustness, accuracy, and time complexity of our algorithm using the ``Robust Vector Alignment Dataset." This dataset contains five template spherical patterns, labeled \(A_1\), \(A_2\), \(A_3\), \(A_4\), and \(A_5\). The patterns are simulated to represent various distributions: Pattern 1 (22212 points) represents a localized simple trajectory, while Pattern 2 (9220 points) represents a more complex trajectory, resembling sharp features observed from objects like drones or other directional vector observations. Patterns 3 (28218 points), 4 (26527 points), and 5 (28557 points) represent spherical distributions covering the entire sphere, capturing characteristics such as random small islands (Pattern 3), large islands (Pattern 4), and non-uniform densities (Pattern 5).

To create the source sets, we add noise and outliers across seven stages (\(B_1\)-\(B_7\)). In Stage 1, there is no noise and are no outliers, meaning the source and template sets contain the same number of points. In Stage 2, Gaussian noise with a standard deviation of 0.01 is added, without any outliers. In Stage 3, we retain the 0.01 Gaussian noise and introduce random outliers by replacing 10\% of the template points. In Stages 4, 5, 6, and 7, we progressively increase the proportion of template points replaced by random outliers to 25\%, 50\%, 75\%, and 90\%, respectively, while maintaining the same noise level as in Stages 1 and 2.

\begin{figure}[ht!]
  \centering
   \includegraphics[width=1.0\linewidth]{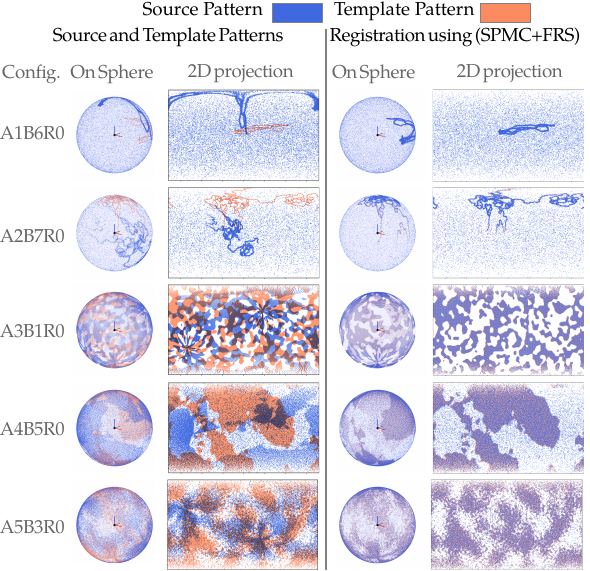}
   \caption{Qualitative results of the SPMC+FRS algorithm on ``Robust Vector Alignment Dataset.'' On the left, the 3D spherical point cloud and 2D projection of the template pattern for each of the five datasets are shown, along with one configuration of the source pattern. On the right, the pattern is displayed after registration. The results demonstrate that the source pattern has successfully aligned with the template pattern. For configuration number, \(A_i\) denotes the template pattern number, \(B_i\) denotes the source set configuration, and \(R_i\) represents the rotation number. }
   \label{fig:sim_qual}
\end{figure}

Unlike previous approaches that select a few specific rotations \cite{makadia2006rotation}, the complete set of source configurations in our experiments is generated by sampling 100 random rotations (R100) over the \(SO(3)\) space (shown in the bottom right of Fig.~\ref*{fig:sim_quant} in the main text) and applying each rotation to the source pattern. Thus, for each template pattern (\(A_1, \ldots, A_5\)), we create a total of \(7 \times 100 = 700\) source patterns. A sample configuration, such as \(A1B7R95\), indicates that the template pattern is \(A_1\), the source pattern is \(B_7\) (which includes 90\% outliers and Gaussian noise with a standard deviation of 0.01), and the rotation applied is labeled as rotation number 95. While \cref{fig:sim_quant} in the main text displays the quantitative results of the simulation experiments, \cref{fig:sim_qual} shows qualitative results of the SPMC+FRS algorithm on several example configurations.

As discussed in Sec.~\ref*{sec:Methods} in the main text, our problem is directly related to the ``Robust Wahba Problem'' \cite{yang2019quaternion}. In their work, Heng Yang et al. developed QUASAR, which achieved state-of-the-art results for the Wahba problem, solving it under extreme outlier conditions (up to 95\% outliers) when putative correspondences are provided. In contrast, our problem assumes no correspondences are given. Therefore, in our first experiments, we compare the performance of QUASAR with our algorithms. Since QUASAR requires putative correspondences, we use FPFH \cite{rusu2009fast} as a prior step before applying QUASAR.


\subsection{Iteration Analysis: }
\label{convergence-analysis}
\begin{figure}[ht!]
  \centering
   \includegraphics[width=1.0\linewidth]{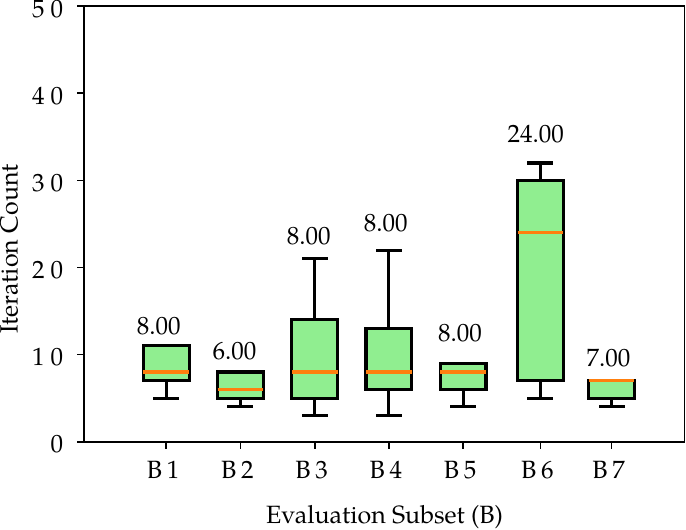}
   \caption{Iteration analysis for the FRS algorithm}
   \label{fig:converge}
\end{figure}

In \cref{fig:converge}, we present the iteration analysis for the FRS algorithm using the ``Robust Vector Alignment Dataset.'' Across all datasets, including the subsets detailed in Sec.~\ref*{simulation_exp} in the main text, the algorithm converges in slightly more than 10 iterations on average.

\section{Experiment 2: Point Cloud Registration}
\label{supplementary-exp-2}

\subsection{Dataset Preparation}
From the ModelNet40 dataset \cite{wu20153d}, we selected 166 objects from 35 object classes. We chose to exclude classes containing highly symmetric objects, such as bowls, cones, vases, bottles, and glass boxes. These objects exhibit symmetry across multiple directions, leading to rotational ambiguity. Even if the algorithm registers these objects accurately, rotational errors may still occur due to their indistinguishable orientations.

\textbf{Complete-to-Complete (Comp2Comp) Dataset}\\
Using the 3D CAD models, we first scaled each model to fit within a unit cube of length 1, then sampled 5000 points per object. We divided the source and target shapes based on two primary criteria:
\textbf{a. No Correspondence} (No corr.): We selected 2500 points for the source and 2500 points for the target, leaving no one-to-one correspondence.
\textbf{b. 10\% Correspondence} (10\% corr.):  In this case, we split the dataset so that the source and target share 10\% of the points with absolute correspondence. We also created two additional cases by adding Gaussian noise with a standard deviation of 0.01 to each configuration. After creating the basic dataset, we applied 10 random rotations and a fixed translation of \([0.1, 0.2, 0.3]\) to simulate the source.

\textbf{Partial-to-Complete (Par2Comp) Dataset}\\
For the Par2Comp dataset, we used the same source dataset but selected 8 equidistant viewpoints across the \(SO(3)\) space to cover the entire sphere. We discarded partial views with ambiguous or highly symmetric content (e.g., one view showing only a portion of an airplane wing where both wings are symmetric). Out of 166 objects, we generated a total of 881 partial views.

\subsection{Centroid Aware Spherical Embedding (CASE)}
\label{supplementary-CASE}

Different spherical projection techniques are discussed in the literature, such as ray tracing from a 3D object from a viewpoint~\cite{cohen2018spherical} and spherical remeshing~\cite{praun2003spherical}. Centroid-Aware Spherical Embedding (CASE) represents the spherical projection of a point cloud relative to its centroid. This embedding is rotation- and scale-invariant and ensures that all points reside on the unit sphere, making it effective for tasks such as point cloud registration.

\begin{figure}[h]
  \centering
   \includegraphics[width=1\linewidth]{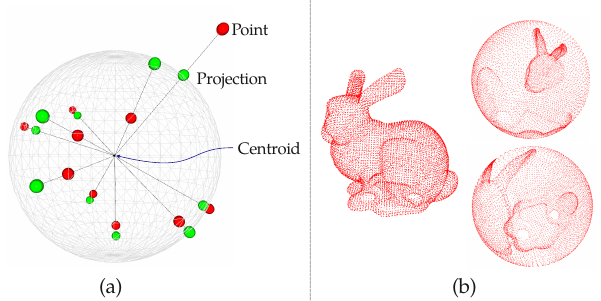}
\caption{CASE Representation of Point Cloud. (a) Point cloud (red) projected onto the unit sphere (green) from the centroid. (b) CASE representation of the bunny point cloud from two different viewpoints.}
   \label{fig:CASE}
\end{figure}

\textbf{Point Cloud to CASE:}
Given a point cloud, let \((x_c, y_c, z_c)\) denote its centroid, and consider a 3D point \((x_1, y_1, z_1)\) in the cloud. The projection of this point onto the unit sphere is computed as follows:

\begin{enumerate}
    \item Compute the vector from the centroid to the point:
    \[
    \vec{v} = \begin{bmatrix} x_1 - x_c \\ y_1 - y_c \\ z_1 - z_c \end{bmatrix}
    \]

    \item Normalize this vector to have a unit norm:
    \[
    \vec{v}_{\text{unit}} = \frac{\vec{v}}{\|\vec{v}\|} = \frac{\begin{bmatrix} x_1 - x_c \\ y_1 - y_c \\ z_1 - z_c \end{bmatrix}}{\sqrt{(x_1 - x_c)^2 + (y_1 - y_c)^2 + (z_1 - z_c)^2}}
    \]

    \item The coordinates of the point projected onto the unit sphere are:
    \[
    \begin{bmatrix} x \\ y \\ z \end{bmatrix} = \vec{v}_{\text{unit}}
    \]
\end{enumerate}

\begin{figure*}[thb]
  \centering
   \includegraphics[width=1\linewidth]{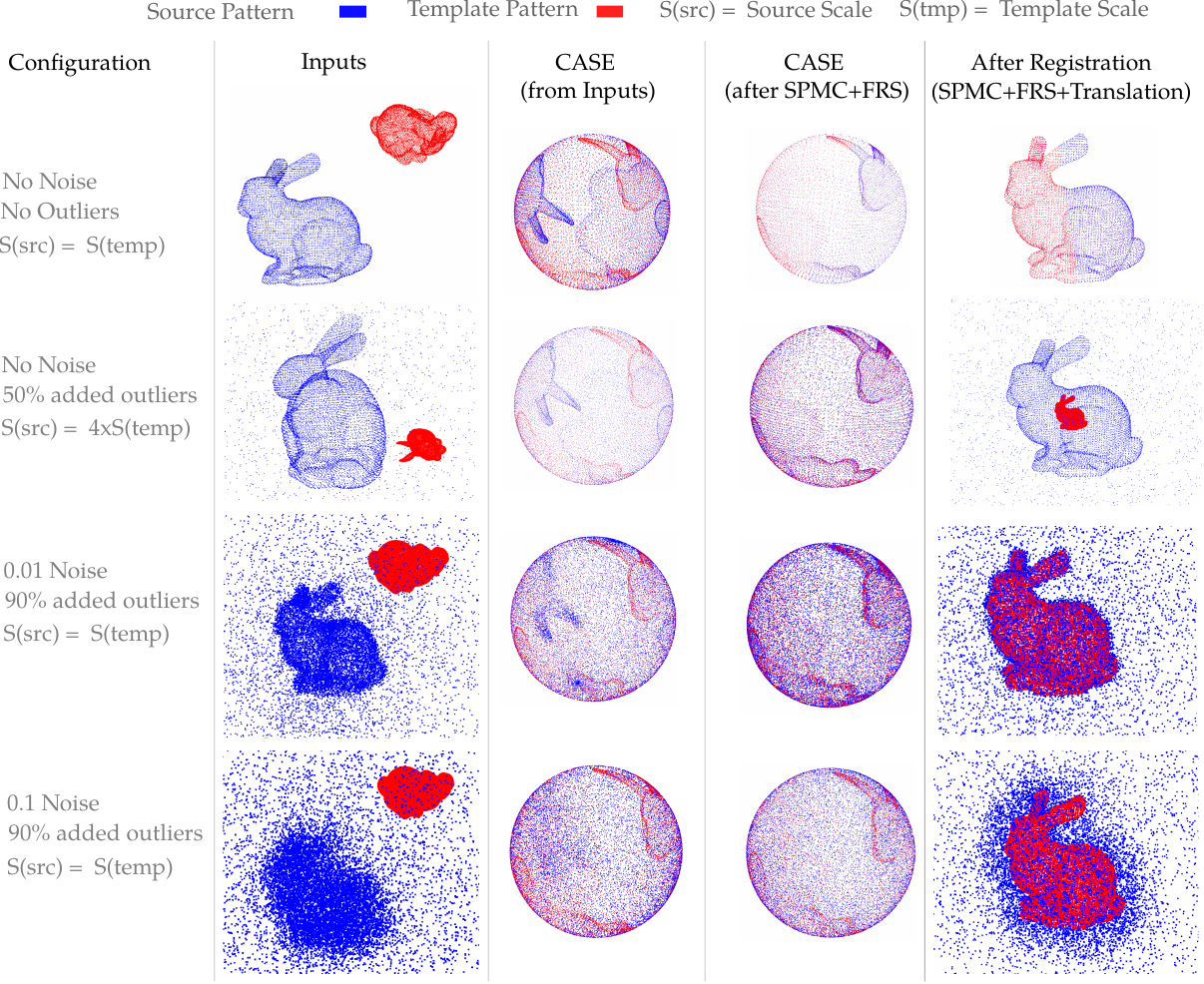}
   \caption{Robustness and Invariance of CASE to Rotation, Translation, and Scale. The first column depicts the input point clouds, where the source set is first scaled to fit into a unit cube. Isotropic Gaussian noise is added to the source set with a mean of 0 and a standard deviation of 0.01 (third row) or 0.1 (fourth row). The second column visualizes the CASE directly derived from the input. The third column shows the CASE after applying SPMC+FRS, and the final column presents the registration results in \(\mathbb{R}^3\) space. The top row illustrates a simple one-to-one case. The second row demonstrates a source set with 50\% outliers and a scale four times that of the template. The third row depicts a source set with the same scale as the template but includes 0.01 Gaussian noise and 90\% added outliers. The bottom row shows a source set with the same scale as the template, 0.1 Gaussian noise, and 90\% added outliers.}
   \label{fig:case_example}
\end{figure*}

\cref{fig:CASE}(a) illustrates a sample point cloud (red) and its projection (green) onto the unit sphere from the centroid. \cref{fig:CASE}(b) depicts the bunny point cloud, highlighting its original structure alongside its CASE representation from two different viewpoints. Additionally, \cref{fig:case_example} demonstrates the robustness of CASE, along with a qualitative evaluation showcasing its invariance to rotation and translation, assuming the centroid is known.

\subsection{Translation Estimation Algorithm}
\label{supplementary-translation-estimation}

 The translation estimation proceeds as follows:

\begin{itemize}
    \item \textbf{Voxel Assignment and Weighting:}  
    Each point in the source and target clouds is assigned to a voxel grid, with voxel size set to 0.2 units. Points are mapped to voxel indices \((i, j, k)\) by applying \(\textit{floor}(\mathbf{p}/\text{voxel\_size})\) for each coordinate \(\mathbf{p}\). A weight is assigned to each voxel, proportional to the number of points it contains, yielding voxel sets \( \mathcal{V}_{\text{source}} \) and \( \mathcal{V}_{\text{target}} \) with respective weights.

    \item \textbf{Translation Vector Voting:}  
    For each voxel pair \((\mathbf{v}_i \in \mathcal{V}_{\text{source}}, \mathbf{u}_j \in \mathcal{V}_{\text{target}})\), a candidate translation vector \(\mathbf{t}_{ij} = (\mathbf{u}_j - \mathbf{v}_i) \times \text{voxel\_size}\) is computed. The vote for each candidate translation \(\mathbf{t}_{ij}\) is weighted by the product of the source and target voxel weights. The translation with the highest accumulated weight across all candidate translations is selected as the coarse translation estimate.

    \item \textbf{Fine Alignment with ICP:}  
    After coarse alignment, the partial source and target clouds will be roughly aligned. For \(N\) points in the partial source cloud, we select \(N\) nearest neighbors from the target cloud and perform translation-only ICP. This refinement iteratively adjusts \(\mathbf{t}_{\text{final}}\) by minimizing residual distances between corresponding points in the aligned source and target clouds.
\end{itemize}

\subsection{Qualitative Results of Point Cloud Registration}

\begin{figure*}[t!]
  \centering
   \includegraphics[width=1\linewidth]{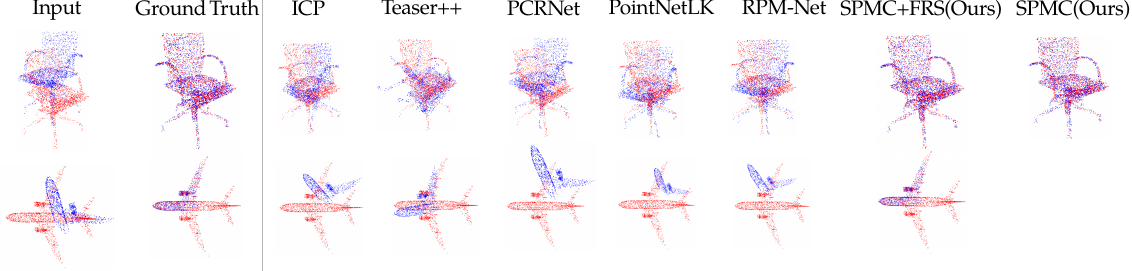}
   \caption{Qualitative results of point cloud registration for complete-to-complete (first row) and partial-to-complete (second row) cases with different methods. These are: ICP \cite{rusinkiewicz2001efficient}, Teaser++ \cite{yang2020teaser}, PCRNet \cite{sarode2019pcrnet}, PointNetLK \cite{aoki2019pointnetlk}, and RPM-Net \cite{yew2020rpm}. For complete-to-complete registration, we used two spherical embeddings: CASE in the SPMC method, and EGI in the SPMC+FRS method. For partial-to-complete registration, only the EGI embedding was used in the SPMC+FRS method.}
   \label{fig:pcr_qual}
\end{figure*}

\cref{fig:pcr_qual} shows qualitative results of our point cloud registration algorithms compared to other works.

\subsection{Point Cloud Registration on Real-World Datasets}
\begin{figure*}[t!]
  \centering
   \includegraphics[width=1\linewidth]{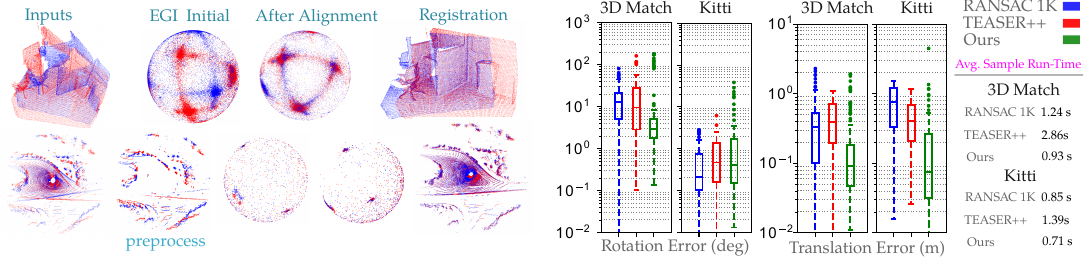}
   \caption{Qualitative (left) and quantitative (right) results on the 3DMatch and KITTI datasets. Point clouds were voxel-downsampled with a voxel size of 0.03. For KITTI, an additional preprocessing step was performed to remove ground reflection points by simply removing the points with very low elevation.}
   \label{fig:pcr_realdatasets}
\end{figure*}

In \cref{fig:pcr_realdatasets}, we evaluate our algorithm on the 3DMatch and KITTI datasets. For 3DMatch, we select scans with an overlap of at least 65\%. Our method is robust when the point clouds have more than 65\% overlap, as is the case for nearly all subsequent scans in KITTI and for many scan pairs in 3DMatch. We exclude cases with less than 65\% overlap, as the method becomes unreliable in such scenarios. A failure case is detailed in \cref{supplementary-translation-Failure-Case}.


We summarize our results to those of other algorithms (EGST \cite{yuan2024egst} and GeDi \cite{poiesi2022learning}) in Table~\ref{tab:comparison}.  EGST \cite{yuan2024egst} appears to be the current state-of-the-art algorithm for point cloud registration, using learned geometric structure descriptors.  GeDi is a recent machine learning method that also uses learned 3D descriptors and was recently state-of-the-art.  The results from these other algorithms were taken from their respective papers.  

Even though our algorithms were not originally intended for point cloud registration, our CASE method with Complete-to-Complete point clouds in Modelnet40 gives a better result than EGST for the ``No correlation and No Noise'' and ``10\% correlation and No Noise'' cases.  Our EGI method gives comparable results to EGST on Modelnet40 when there is Gaussian noise and for partial point clouds.  However, their algorithm is much more accurate on the KITTI and 3DMatch datasets.  The GeDi algorithm is comparable to ours on KITTI and 3DMatch.  

The EGST algorithm paper quotes an execution time of 0.012~s on the partial overlap data, as compared to 0.05~s for ours (although they used a faster computer).  The GeDi algorithm is much slower than ours, since their algorithm is slightly faster than FPFH, and ours is roughly 20x faster than FPFH.

\begin{table*}[ht]
\centering
\caption{Rotation (\(E(R)\), [°]) and Translation (\(E(t)\), [cm]) Errors across different datasets and algorithms.}
\label{tab:comparison}
\begin{tabular}{llcc}
\toprule
\textbf{Dataset/Scenario} & \textbf{Method} & \(E(R)\) [°] & \(E(t)\) [cm]  \\ \midrule
\multicolumn{4}{l}{\textbf{Modelnet40: Complete-Complete}} \\
Unseen Objects    & EGST HA \cite{yuan2024egst} & 0.1803° & 0.15    \\
Unseen Categories & EGST HA \cite{yuan2024egst} & 0.4554° & 0.32    \\
Gaussian Noise    & EGST HA \cite{yuan2024egst} & 1.1687° & 0.96    \\
Comp2Comp: No corr., No Noise   & Ours - CASE  & 0.029°  & 0.13      \\ 
Comp2Comp: No corr., 0.01 Noise & Ours - CASE  & 1.1°  & 0.8     \\
Comp2Comp: 10\% corr., No Noise & Ours - CASE  & 0.001°  & 0.1      \\
Comp2Comp: 10\% corr., 0.01 Noise & Ours - CASE & 1.7°  & 0.8      \\ 

Comp2Comp: No corr., No Noise   & Ours - EGI  & 1.3°  & 0.8      \\ 
Comp2Comp: No corr., 0.01 Noise & Ours - EGI  & 1.7°  & 0.8     \\
Comp2Comp: 10\% corr., No Noise & Ours - EGI  & 1.6°  & 0.8      \\
Comp2Comp: 10\% corr., 0.01 Noise & Ours - EGI & 1.7°  & 0.8      \\ \midrule

\multicolumn{4}{l}{\textbf{Modelnet40: Partial-Partial}} \\
Partial-Partial   & EGST HA \cite{yuan2024egst} & 3.3040° & 4.92    \\   
Part2Comp: No corr., No Noise & Ours - EGI   & 3.2°   & 3.9   \\
Part2Comp: No corr., 0.01 Noise & Ours - EGI   & 3.0°   & 4.2   \\
Part2Comp: 10\% corr., No Noise & Ours - EGI   & 2.3°   &  7.8   \\
Part2Comp: 10\% corr., 0.01 Noise & Ours - EGI   & 3.0°   & 5.8   \\ \midrule

\multicolumn{4}{l}{\textbf{KITTI}} \\
KITTI             & EGST \cite{yuan2024egst}    & 0.0168° & 0.18   \\

3DMatch (training) \(\rightarrow\) KITTI (testing) & GeDi \cite{poiesi2022learning} & 0.40°  & 8.21--10.34  \\
KITTI (training) \(\rightarrow\) KITTI (testing) & GeDi \cite{poiesi2022learning} & 0.32--0.33° & 7.22--7.55     \\ 
KITTI            & Ours                 & 0.42° & 7.5    \\ \midrule
\multicolumn{4}{l}{\textbf{3DMatch}} \\ 
3DMatch           & EGST \cite{yuan2024egst}   & 0.2086° & 0.87   \\ 
3DMatch          & Ours                 & 3.0°   & 9.2      \\ \bottomrule
\end{tabular}
\end{table*}

\subsection{Failure Case}
\label{supplementary-translation-Failure-Case}

\begin{figure*}[t!]
  \centering
   \includegraphics[width=1\linewidth]{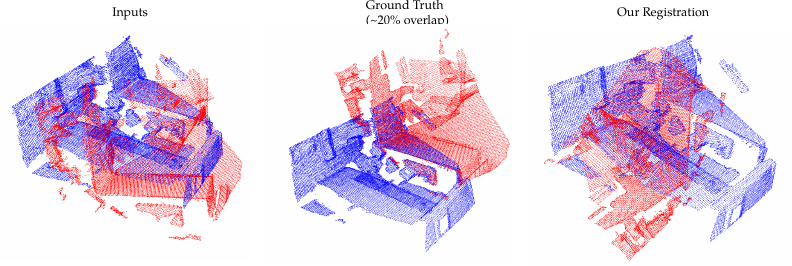}
   \caption{Example of point cloud registration where our algorithm fails. In this example, there is only around 20\% overlap between the two input point clouds, leading the algorithm to not work correctly.}
   \label{fig:pcr_failure}
\end{figure*}

\cref{fig:pcr_failure} shows an example of how our algorithm can fail with point cloud registration if there is insufficient overlap between the two point clouds.  In the example, there is only \textasciitilde20\% overlap; the algorithm works reliably if there is $>$65\% overlap.

\section{Experiment 3: Rotation Estimation From Spherical Images}
\label{supplementary-exp-3}
\begin{figure*}[p!]
  \centering
   \includegraphics[width=0.55\linewidth]{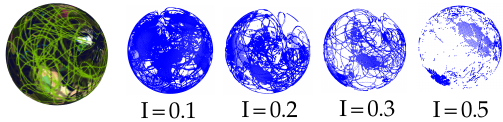}
   \caption{Effect of SphImg2SphPoints at different threshold intensities.}
   \label{fig:sph_intensity}
\end{figure*}

While converting spherical images to spherical points (see Sec.~\ref*{experiment3} in the main text), the choice of threshold value is crucial. A lower threshold captures more features but includes more clutter, while a higher threshold captures fewer features with reduced clutter. \cref{fig:sph_intensity} illustrates the effects of varying threshold values.

\subsection{Qualitative Results}

\begin{figure*}[t!]
  \centering
   \includegraphics[width=1\linewidth]{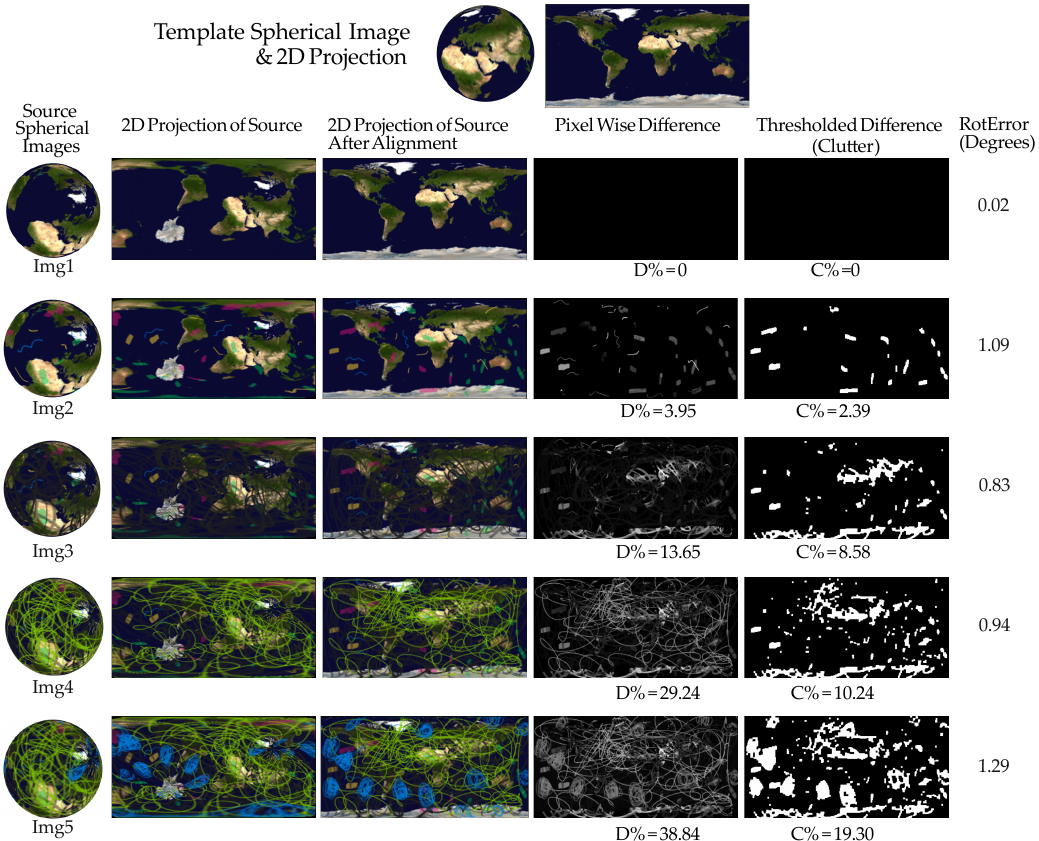}
   \caption{Results of rotation estimation from spherical images. The top row displays the template spherical image and its 2D projection. In the next six consecutive rows, source spherical images undergo varying levels of pixelwise difference (D\%) or thresholded difference/clutter (C\%). For this case, the input rotations were XYZ Euler angles with \(\alpha = 55.27^\circ\), \(\beta = 12.11^\circ\), and \(\gamma = 11.02^\circ\). The estimated rotation error using our algorithm is shown in the right columns.}
   \label{fig:sph_result}
\end{figure*}

\cref{fig:sph_result} shows all source spherical images and their 2D projections alongside the template spherical image and its 2D projection for a sample rotation. The results after alignment using our algorithm are also displayed, including a pixel-wise difference map and a binary thresholded map (clutter map) for each configuration.

%


%% file: main.bbl
\begin{thebibliography}{82}
\providecommand{\natexlab}[1]{#1}
\providecommand{\url}[1]{\texttt{#1}}
\expandafter\ifx\csname urlstyle\endcsname\relax
  \providecommand{\doi}[1]{doi: #1}\else
  \providecommand{\doi}{doi: \begingroup \urlstyle{rm}\Url}\fi

\bibitem[Antonante et~al.(2021)Antonante, Tzoumas, Yang, and Carlone]{antonante2021outlier}
Pasquale Antonante, Vasileios Tzoumas, Heng Yang, and Luca Carlone.
\newblock Outlier-robust estimation: Hardness, minimally tuned algorithms, and applications.
\newblock \emph{IEEE Transactions on Robotics}, 38\penalty0 (1):\penalty0 281--301, 2021.

\bibitem[Aoki et~al.(2019)Aoki, Goforth, Srivatsan, and Lucey]{aoki2019pointnetlk}
Yasuhiro Aoki, Hunter Goforth, Rangaprasad~Arun Srivatsan, and Simon Lucey.
\newblock Pointnetlk: Robust \& efficient point cloud registration using pointnet.
\newblock In \emph{Proceedings of the IEEE/CVF conference on computer vision and pattern recognition}, pages 7163--7172, 2019.

\bibitem[Arun et~al.(1987)Arun, Huang, and Blostein]{arun1987least}
K~Somani Arun, Thomas~S Huang, and Steven~D Blostein.
\newblock Least-squares fitting of two 3-d point sets.
\newblock \emph{IEEE Transactions on pattern analysis and machine intelligence}, PAMI-9\penalty0 (5):\penalty0 698--700, 1987.

\bibitem[Bai et~al.(2021)Bai, Luo, Zhou, Chen, Li, Hu, Fu, and Tai]{bai2021pointdsc}
Xuyang Bai, Zixin Luo, Lei Zhou, Hongkai Chen, Lei Li, Zeyu Hu, Hongbo Fu, and Chiew-Lan Tai.
\newblock Pointdsc: Robust point cloud registration using deep spatial consistency.
\newblock In \emph{Proceedings of the IEEE/CVF Conference on Computer Vision and Pattern Recognition}, pages 15859--15869, 2021.

\bibitem[Bauer et~al.(2021)Bauer, Patten, and Vincze]{bauer2021reagent}
Dominik Bauer, Timothy Patten, and Markus Vincze.
\newblock Reagent: Point cloud registration using imitation and reinforcement learning.
\newblock In \emph{Proceedings of the IEEE/CVF Conference on Computer Vision and Pattern Recognition}, pages 14586--14594, 2021.

\bibitem[Bazin et~al.(2014)Bazin, Seo, Hartley, and Pollefeys]{bazin2014globally}
Jean-Charles Bazin, Yongduek Seo, Richard Hartley, and Marc Pollefeys.
\newblock Globally optimal inlier set maximization with unknown rotation and focal length.
\newblock In \emph{Computer Vision--ECCV 2014: 13th European Conference, Zurich, Switzerland, September 6-12, 2014, Proceedings, Part II 13}, pages 803--817. Springer, 2014.

\bibitem[Bernreiter et~al.(2021)Bernreiter, Ott, Nieto, Siegwart, and Cadena]{bernreiter2021phaser}
Lukas Bernreiter, Lionel Ott, Juan Nieto, Roland Siegwart, and Cesar Cadena.
\newblock Phaser: A robust and correspondence-free global pointcloud registration.
\newblock \emph{IEEE Robotics and Automation Letters}, 6\penalty0 (2):\penalty0 855--862, 2021.

\bibitem[Besl and McKay(1992)]{besl1992method}
Paul~J Besl and Neil~D McKay.
\newblock Method for registration of 3-d shapes.
\newblock In \emph{Sensor fusion IV: control paradigms and data structures}, pages 586--606. Spie, 1992.

\bibitem[Blais and Levine(1995)]{blais1995registering}
G{\'e}rard Blais and Martin~D. Levine.
\newblock Registering multiview range data to create 3d computer objects.
\newblock \emph{IEEE Transactions on Pattern Analysis and Machine Intelligence}, 17\penalty0 (8):\penalty0 820--824, 1995.

\bibitem[Bustos and Chin(2017)]{bustos2017guaranteed}
Alvaro~Parra Bustos and Tat-Jun Chin.
\newblock Guaranteed outlier removal for point cloud registration with correspondences.
\newblock \emph{IEEE transactions on pattern analysis and machine intelligence}, 40\penalty0 (12):\penalty0 2868--2882, 2017.

\bibitem[Campbell and Petersson(2016)]{campbell2016gogma}
Dylan Campbell and Lars Petersson.
\newblock Gogma: Globally-optimal gaussian mixture alignment.
\newblock In \emph{Proceedings of the IEEE conference on computer vision and pattern recognition}, pages 5685--5694, 2016.

\bibitem[Carpentier and Einbond(2023)]{carpentier2023spherical}
Thibaut Carpentier and Aaron Einbond.
\newblock Spherical correlation as a similarity measure for 3-d radiation patterns of musical instruments.
\newblock \emph{Acta Acustica}, 7:\penalty0 40, 2023.

\bibitem[Cheng and Crassidis(2019)]{cheng2019total}
Yang Cheng and John~L Crassidis.
\newblock A total least-squares estimate for attitude determination.
\newblock In \emph{AIAA Scitech 2019 Forum}, page 1176, 2019.

\bibitem[Chetverikov et~al.(2002)Chetverikov, Svirko, Stepanov, and Krsek]{chetverikov2002trimmed}
Dmitry Chetverikov, Dmitry Svirko, Dmitry Stepanov, and Pavel Krsek.
\newblock The trimmed iterative closest point algorithm.
\newblock In \emph{2002 International Conference on Pattern Recognition}, pages 545--548. IEEE, 2002.

\bibitem[Chin et~al.(2016)Chin, Heng~Kee, Eriksson, and Neumann]{chin2016guaranteed}
Tat-Jun Chin, Yang Heng~Kee, Anders Eriksson, and Frank Neumann.
\newblock Guaranteed outlier removal with mixed integer linear programs.
\newblock In \emph{Proceedings of the IEEE Conference on Computer Vision and Pattern Recognition}, pages 5858--5866, 2016.

\bibitem[Chin et~al.(2019)Chin, Bagchi, Eriksson, and Van~Schaik]{chin2019star}
Tat-Jun Chin, Samya Bagchi, Anders Eriksson, and Andre Van~Schaik.
\newblock Star tracking using an event camera.
\newblock In \emph{Proceedings of the IEEE/CVF Conference on Computer Vision and Pattern Recognition Workshops}, pages 0--0, 2019.

\bibitem[Choy et~al.(2019)Choy, Park, and Koltun]{choy2019fully}
Christopher Choy, Jaesik Park, and Vladlen Koltun.
\newblock Fully convolutional geometric features.
\newblock In \emph{Proceedings of the IEEE/CVF international conference on computer vision}, pages 8958--8966, 2019.

\bibitem[Choy et~al.(2020)Choy, Dong, and Koltun]{choy2020deep}
Christopher Choy, Wei Dong, and Vladlen Koltun.
\newblock Deep global registration.
\newblock In \emph{Proceedings of the IEEE/CVF conference on computer vision and pattern recognition}, pages 2514--2523, 2020.

\bibitem[Chui and Rangarajan(2003)]{chui2003new}
Haili Chui and Anand Rangarajan.
\newblock A new point matching algorithm for non-rigid registration.
\newblock \emph{Computer Vision and Image Understanding}, 89\penalty0 (2-3):\penalty0 114--141, 2003.

\bibitem[Cohen et~al.(2018)Cohen, Geiger, K{\"o}hler, and Welling]{cohen2018spherical}
Taco~S Cohen, Mario Geiger, Jonas K{\"o}hler, and Max Welling.
\newblock Spherical cnns.
\newblock \emph{arXiv preprint arXiv:1801.10130}, 2018.

\bibitem[Drost et~al.(2010)Drost, Ulrich, Navab, and Ilic]{drost2010model}
Bertram Drost, Markus Ulrich, Nassir Navab, and Slobodan Ilic.
\newblock Model globally, match locally: Efficient and robust 3d object recognition.
\newblock In \emph{2010 IEEE computer society conference on computer vision and pattern recognition}, pages 998--1005. Ieee, 2010.

\bibitem[Esteves et~al.(2023)Esteves, Slotine, and Makadia]{esteves2023scaling}
Carlos Esteves, Jean-Jacques Slotine, and Ameesh Makadia.
\newblock Scaling spherical cnns.
\newblock \emph{arXiv preprint arXiv:2306.05420}, 2023.

\bibitem[Fischler and Bolles(1981)]{fischler1981random}
Martin~A Fischler and Robert~C Bolles.
\newblock Random sample consensus: a paradigm for model fitting with applications to image analysis and automated cartography.
\newblock \emph{Communications of the ACM}, 24\penalty0 (6):\penalty0 381--395, 1981.

\bibitem[Forbes and de~Ruiter(2015)]{forbes2015linear}
James~Richard Forbes and Anton~HJ de Ruiter.
\newblock Linear-matrix-inequality-based solution to wahba’s problem.
\newblock \emph{Journal of guidance, control, and dynamics}, 38\penalty0 (1):\penalty0 147--151, 2015.

\bibitem[Geiger et~al.(2013)Geiger, Lenz, Stiller, and Urtasun]{geiger2013vision}
Andreas Geiger, Philip Lenz, Christoph Stiller, and Raquel Urtasun.
\newblock Vision meets robotics: The kitti dataset.
\newblock \emph{The international journal of robotics research}, 32\penalty0 (11):\penalty0 1231--1237, 2013.

\bibitem[Gower and Dijksterhuis(2004)]{gower2004procrustes}
John~C Gower and Garmt~B Dijksterhuis.
\newblock \emph{Procrustes problems}.
\newblock OUP Oxford, 2004.

\bibitem[Gutman et~al.(2008)Gutman, Wang, Chan, Thompson, and Toga]{gutman2008shape}
Boris Gutman, Yalin Wang, Tony Chan, Paul~M Thompson, and Arthur~W Toga.
\newblock Shape registration with spherical cross correlation.
\newblock In \emph{2nd MICCAI Workshop on Mathematical Foundations of Computational Anatomy}, pages 56--67, 2008.

\bibitem[Hartley and Kahl(2009)]{hartley2009global}
Richard~I Hartley and Fredrik Kahl.
\newblock Global optimization through rotation space search.
\newblock \emph{International Journal of Computer Vision}, 82\penalty0 (1):\penalty0 64--79, 2009.

\bibitem[Horn(1984)]{EGIHorn}
B.K.P. Horn.
\newblock Extended gaussian images.
\newblock \emph{Proceedings of the IEEE}, 72\penalty0 (12):\penalty0 1671--1686, 1984.

\bibitem[Horn(1987)]{horn1987closed}
Berthold~KP Horn.
\newblock Closed-form solution of absolute orientation using unit quaternions.
\newblock \emph{Josa a}, 4\penalty0 (4):\penalty0 629--642, 1987.

\bibitem[Horn et~al.(1988)Horn, Hilden, and Negahdaripour]{horn1988closed}
Berthold~KP Horn, Hugh~M Hilden, and Shahriar Negahdaripour.
\newblock Closed-form solution of absolute orientation using orthonormal matrices.
\newblock \emph{Josa a}, 5\penalty0 (7):\penalty0 1127--1135, 1988.

\bibitem[Huang et~al.(2021)Huang, Gojcic, Usvyatsov, Wieser, and Schindler]{huang2021predator}
Shengyu Huang, Zan Gojcic, Mikhail Usvyatsov, Andreas Wieser, and Konrad Schindler.
\newblock Predator: Registration of 3d point clouds with low overlap.
\newblock In \emph{Proceedings of the IEEE/CVF Conference on computer vision and pattern recognition}, pages 4267--4276, 2021.

\bibitem[Jian and Vemuri(2010)]{jian2010robust}
Bing Jian and Baba~C Vemuri.
\newblock Robust point set registration using gaussian mixture models.
\newblock \emph{IEEE transactions on pattern analysis and machine intelligence}, 33\penalty0 (8):\penalty0 1633--1645, 2010.

\bibitem[Kabsch(1976)]{kabsch1976solution}
Wolfgang Kabsch.
\newblock A solution for the best rotation to relate two sets of vectors.
\newblock \emph{Acta Crystallographica Section A: Crystal Physics, Diffraction, Theoretical and General Crystallography}, 32\penalty0 (5):\penalty0 922--923, 1976.

\bibitem[Khoshelham(2016)]{khoshelham2016closed}
Kourosh Khoshelham.
\newblock Closed-form solutions for estimating a rigid motion from plane correspondences extracted from point clouds.
\newblock \emph{ISPRS Journal of Photogrammetry and Remote Sensing}, 114:\penalty0 78--91, 2016.

\bibitem[Laguna et~al.(1994)]{laguna1994time}
Pablo Laguna et~al.
\newblock {A time delay estimator based on the signal integral: Theoretical performance and testing on ECG signals}.
\newblock \emph{IEEE Trans. on Sig. Proc.}, 42\penalty0 (11):\penalty0 3224--3229, 1994.

\bibitem[Ley and Verdebout(2017)]{ley2017modern}
Christophe Ley and Thomas Verdebout.
\newblock \emph{Modern directional statistics}.
\newblock Chapman and Hall/CRC, 2017.

\bibitem[Li et~al.(2011)Li, Mordukhovich, Wang, and Yao]{li2011weak}
Chong Li, Boris~S Mordukhovich, Jinhua Wang, and Jen-Chih Yao.
\newblock Weak sharp minima on riemannian manifolds.
\newblock \emph{SIAM Journal on Optimization}, 21\penalty0 (4):\penalty0 1523--1560, 2011.

\bibitem[Li and Hartley(2007)]{li20073d}
Hongdong Li and Richard Hartley.
\newblock The 3d-3d registration problem revisited.
\newblock In \emph{2007 IEEE 11th international conference on computer vision}, pages 1--8. IEEE, 2007.

\bibitem[Li et~al.(2020)Li, Hu, and Ai]{li2020gesac}
Jiayuan Li, Qingwu Hu, and Mingyao Ai.
\newblock Gesac: Robust graph enhanced sample consensus for point cloud registration.
\newblock \emph{ISPRS Journal of Photogrammetry and Remote Sensing}, 167:\penalty0 363--374, 2020.

\bibitem[Li et~al.(2021)Li, Hu, and Ai]{li2021point}
Jiayuan Li, Qingwu Hu, and Mingyao Ai.
\newblock Point cloud registration based on one-point ransac and scale-annealing biweight estimation.
\newblock \emph{IEEE Transactions on Geoscience and Remote Sensing}, 59\penalty0 (11):\penalty0 9716--9729, 2021.

\bibitem[Lian et~al.(2016)Lian, Zhang, and Yang]{lian2016efficient}
Wei Lian, Lei Zhang, and Ming-Hsuan Yang.
\newblock An efficient globally optimal algorithm for asymmetric point matching.
\newblock \emph{IEEE transactions on pattern analysis and machine intelligence}, 39\penalty0 (7):\penalty0 1281--1293, 2016.

\bibitem[Lin et~al.(2021)Lin, Chung, and Wang]{lin2021self}
Huei-Yung Lin, Yuan-Chi Chung, and Ming-Liang Wang.
\newblock Self-localization of mobile robots using a single catadioptric camera with line feature extraction.
\newblock \emph{Sensors}, 21\penalty0 (14):\penalty0 4719, 2021.

\bibitem[Liu et~al.(2018)Liu, Wang, Song, and Wang]{liu2018efficient}
Yinlong Liu, Chen Wang, Zhijian Song, and Manning Wang.
\newblock Efficient global point cloud registration by matching rotation invariant features through translation search.
\newblock In \emph{Proceedings of the European Conference on Computer Vision (ECCV)}, pages 448--463, 2018.

\bibitem[Ma et~al.(2016)Ma, Guo, Zhao, Lu, Zhang, and Wan]{ma2016fast}
Yanxin Ma, Yulan Guo, Jian Zhao, Min Lu, Jun Zhang, and Jianwei Wan.
\newblock Fast and accurate registration of structured point clouds with small overlaps.
\newblock In \emph{Proceedings of the IEEE Conference on Computer Vision and Pattern Recognition Workshops}, pages 1--9, 2016.

\bibitem[Makadia and Daniilidis(2003)]{makadia2003direct}
Ameesh Makadia and Kostas Daniilidis.
\newblock Direct 3d-rotation estimation from spherical images via a generalized shift theorem.
\newblock In \emph{2003 IEEE Computer Society Conference on Computer Vision and Pattern Recognition, 2003. Proceedings.}, pages II--217. IEEE, 2003.

\bibitem[Makadia and Daniilidis(2006)]{makadia2006rotation}
Ameesh Makadia and Kostas Daniilidis.
\newblock Rotation recovery from spherical images without correspondences.
\newblock \emph{IEEE transactions on pattern analysis and machine intelligence}, 28\penalty0 (7):\penalty0 1170--1175, 2006.

\bibitem[Makadia et~al.(2004)Makadia, Sorgi, and Daniilidis]{makadia2004rotation}
Ameesh Makadia, Lorenzo Sorgi, and Kostas Daniilidis.
\newblock Rotation estimation from spherical images.
\newblock In \emph{Proceedings of the 17th International Conference on Pattern Recognition, 2004. ICPR 2004.}, pages 590--593. IEEE, 2004.

\bibitem[Makadia et~al.(2006)Makadia, Patterson, and Daniilidis]{makadia2006fully}
Ameesh Makadia, Alexander Patterson, and Kostas Daniilidis.
\newblock Fully automatic registration of 3d point clouds.
\newblock In \emph{2006 IEEE Computer Society Conference on Computer Vision and Pattern Recognition (CVPR'06)}, pages 1297--1304. IEEE, 2006.

\bibitem[Mardia and Jupp(2009)]{mardia2009directional}
Kanti~V Mardia and Peter~E Jupp.
\newblock \emph{Directional statistics}.
\newblock John Wiley \& Sons, 2009.

\bibitem[Markley(1988)]{markley1988attitude}
F~Landis Markley.
\newblock Attitude determination using vector observations and the singular value decomposition.
\newblock \emph{Journal of the Astronautical Sciences}, 36\penalty0 (3):\penalty0 245--258, 1988.

\bibitem[Myronenko and Song(2010)]{myronenko2010point}
Andriy Myronenko and Xubo Song.
\newblock Point set registration: Coherent point drift.
\newblock \emph{IEEE transactions on pattern analysis and machine intelligence}, 32\penalty0 (12):\penalty0 2262--2275, 2010.

\bibitem[Parra et~al.(2019)Parra, Chin, Neumann, Friedrich, and Katzmann]{parra2019practical}
Alvaro Parra, Tat-Jun Chin, Frank Neumann, Tobias Friedrich, and Maximilian Katzmann.
\newblock A practical maximum clique algorithm for matching with pairwise constraints.
\newblock \emph{arXiv preprint arXiv:1902.01534}, 2019.

\bibitem[Parra~Bustos and Chin(2015)]{parra2015guaranteed}
Alvaro Parra~Bustos and Tat-Jun Chin.
\newblock Guaranteed outlier removal for rotation search.
\newblock In \emph{Proceedings of the IEEE International Conference on Computer Vision}, pages 2165--2173, 2015.

\bibitem[Parra~Bustos et~al.(2014)Parra~Bustos, Chin, and Suter]{parra2014fast}
Alvaro Parra~Bustos, Tat-Jun Chin, and David Suter.
\newblock Fast rotation search with stereographic projections for 3d registration.
\newblock In \emph{Proceedings of the IEEE conference on computer vision and pattern recognition}, pages 3930--3937, 2014.

\bibitem[Peng et~al.(2022)Peng, Tsakiris, and Vidal]{peng2022arcs}
Liangzu Peng, Manolis~C Tsakiris, and Ren{\'e} Vidal.
\newblock Arcs: Accurate rotation and correspondence search.
\newblock In \emph{Proceedings of the IEEE/CVF Conference on Computer Vision and Pattern Recognition}, pages 11153--11163, 2022.

\bibitem[Poiesi and Boscaini(2022)]{poiesi2022learning}
Fabio Poiesi and Davide Boscaini.
\newblock Learning general and distinctive 3d local deep descriptors for point cloud registration.
\newblock \emph{IEEE TPAMI}, 45\penalty0 (3):\penalty0 3979--3985, 2022.

\bibitem[Praun and Hoppe(2003)]{praun2003spherical}
Emil Praun and Hugues Hoppe.
\newblock Spherical parametrization and remeshing.
\newblock \emph{ACM transactions on graphics (TOG)}, 22\penalty0 (3):\penalty0 340--349, 2003.

\bibitem[Rusinkiewicz and Levoy(2001)]{rusinkiewicz2001efficient}
Szymon Rusinkiewicz and Marc Levoy.
\newblock Efficient variants of the icp algorithm.
\newblock In \emph{Proceedings third international conference on 3-D digital imaging and modeling}, pages 145--152. IEEE, 2001.

\bibitem[Rusu et~al.(2009)Rusu, Blodow, and Beetz]{rusu2009fast}
Radu~Bogdan Rusu, Nico Blodow, and Michael Beetz.
\newblock Fast point feature histograms (fpfh) for 3d registration.
\newblock In \emph{2009 IEEE international conference on robotics and automation}, pages 3212--3217. IEEE, 2009.

\bibitem[Saini(2020)]{learning3d}
Vinit Saini.
\newblock {Learning3D: A Modern Library for Deep Learning on 3D Point Clouds Data}.
\newblock \url{https://github.com/vinits5/learning3d}, 2020.
\newblock Accessed: 2024-11-12.

\bibitem[Sarode et~al.(2019)Sarode, Li, Goforth, Aoki, Srivatsan, Lucey, and Choset]{sarode2019pcrnet}
Vinit Sarode, Xueqian Li, Hunter Goforth, Yasuhiro Aoki, Rangaprasad~Arun Srivatsan, Simon Lucey, and Howie Choset.
\newblock Pcrnet: Point cloud registration network using pointnet encoding.
\newblock \emph{arXiv preprint arXiv:1908.07906}, 2019.

\bibitem[Saunderson et~al.(2015)Saunderson, Parrilo, and Willsky]{saunderson2015semidefinite}
James Saunderson, Pablo~A Parrilo, and Alan~S Willsky.
\newblock Semidefinite descriptions of the convex hull of rotation matrices.
\newblock \emph{SIAM Journal on Optimization}, 25\penalty0 (3):\penalty0 1314--1343, 2015.

\bibitem[Sch{\"o}nemann(1966)]{schonemann1966generalized}
Peter~H Sch{\"o}nemann.
\newblock A generalized solution of the orthogonal procrustes problem.
\newblock \emph{Psychometrika}, 31\penalty0 (1):\penalty0 1--10, 1966.

\bibitem[Shi et~al.(2021)Shi, Yang, and Carlone]{shi2021robin}
Jingnan Shi, Heng Yang, and Luca Carlone.
\newblock Robin: a graph-theoretic approach to reject outliers in robust estimation using invariants.
\newblock In \emph{2021 IEEE International Conference on Robotics and Automation (ICRA)}, pages 13820--13827. IEEE, 2021.

\bibitem[Sorgi and Daniilidis(2004)]{sorgi2004normalized}
Lorenzo Sorgi and Kostas Daniilidis.
\newblock Normalized cross-correlation for spherical images.
\newblock In \emph{Computer Vision-ECCV 2004: 8th European Conference on Computer Vision, Prague, Czech Republic, May 11-14, 2004. Proceedings, Part II 8}, pages 542--553. Springer, 2004.

\bibitem[Straub et~al.(2017)Straub, Campbell, How, and Fisher]{straub2017efficient}
Julian Straub, Trevor Campbell, Jonathan~P How, and John~W Fisher.
\newblock Efficient global point cloud alignment using bayesian nonparametric mixtures.
\newblock In \emph{Proceedings of the IEEE Conference on Computer Vision and Pattern Recognition}, pages 2941--2950, 2017.

\bibitem[Sun(2021)]{sun2021ransic}
Lei Sun.
\newblock Ransic: Fast and highly robust estimation for rotation search and point cloud registration using invariant compatibility.
\newblock \emph{IEEE Robotics and Automation Letters}, 7\penalty0 (1):\penalty0 143--150, 2021.

\bibitem[Talak et~al.(2023)Talak, Peng, and Carlone]{Talak23}
Rajat Talak, Lisa~R. Peng, and Luca Carlone.
\newblock Certifiable object pose estimation: Foundations, learning models, and self-training.
\newblock \emph{IEEE Transactions on Robotics}, 39\penalty0 (4):\penalty0 2805--2824, 2023.

\bibitem[Tombari et~al.(2013)Tombari, Salti, and Di~Stefano]{tombari2013performance}
Federico Tombari, Samuele Salti, and Luigi Di~Stefano.
\newblock Performance evaluation of 3d keypoint detectors.
\newblock \emph{International Journal of Computer Vision}, 102\penalty0 (1):\penalty0 198--220, 2013.

\bibitem[Wahba(1965)]{wahba1965least}
Grace Wahba.
\newblock A least squares estimate of satellite attitude.
\newblock \emph{SIAM review}, 7\penalty0 (3):\penalty0 409--409, 1965.

\bibitem[Wu et~al.(2015)Wu, Song, Khosla, Yu, Zhang, Tang, and Xiao]{wu20153d}
Zhirong Wu, Shuran Song, Aditya Khosla, Fisher Yu, Linguang Zhang, Xiaoou Tang, and Jianxiong Xiao.
\newblock 3d shapenets: A deep representation for volumetric shapes.
\newblock In \emph{Proceedings of the IEEE conference on computer vision and pattern recognition}, pages 1912--1920, 2015.

\bibitem[Yang and Carlone(2019)]{yang2019quaternion}
Heng Yang and Luca Carlone.
\newblock A quaternion-based certifiably optimal solution to the wahba problem with outliers.
\newblock In \emph{Proceedings of the IEEE/CVF International Conference on Computer Vision}, pages 1665--1674, 2019.

\bibitem[Yang et~al.(2020{\natexlab{a}})Yang, Antonante, Tzoumas, and Carlone]{yang2020graduated}
Heng Yang, Pasquale Antonante, Vasileios Tzoumas, and Luca Carlone.
\newblock Graduated non-convexity for robust spatial perception: From non-minimal solvers to global outlier rejection.
\newblock \emph{IEEE Robotics and Automation Letters}, 5\penalty0 (2):\penalty0 1127--1134, 2020{\natexlab{a}}.

\bibitem[Yang et~al.(2020{\natexlab{b}})Yang, Shi, and Carlone]{yang2020teaser}
Heng Yang, Jingnan Shi, and Luca Carlone.
\newblock Teaser: Fast and certifiable point cloud registration.
\newblock \emph{IEEE Transactions on Robotics}, 37\penalty0 (2):\penalty0 314--333, 2020{\natexlab{b}}.

\bibitem[Yang et~al.(2015)Yang, Li, Campbell, and Jia]{yang2015go}
Jiaolong Yang, Hongdong Li, Dylan Campbell, and Yunde Jia.
\newblock Go-icp: A globally optimal solution to 3d icp point-set registration.
\newblock \emph{IEEE transactions on pattern analysis and machine intelligence}, 38\penalty0 (11):\penalty0 2241--2254, 2015.

\bibitem[Yew and Lee(2020)]{yew2020rpm}
Zi~Jian Yew and Gim~Hee Lee.
\newblock Rpm-net: Robust point matching using learned features.
\newblock In \emph{Proceedings of the IEEE/CVF conference on computer vision and pattern recognition}, pages 11824--11833, 2020.

\bibitem[Yuan et~al.(2024)]{yuan2024egst}
Yongzhe Yuan et~al.
\newblock {EGST: Enhanced geometric structure transformer for point cloud registration}.
\newblock \emph{{IEEE Trans. on Vis. and Comp. Graphics}}, 30\penalty0 (9):\penalty0 6222--6234, 2024.

\bibitem[Zeng et~al.(2017)Zeng, Song, Nie{\ss}ner, Fisher, Xiao, and Funkhouser]{zeng20173dmatch}
Andy Zeng, Shuran Song, Matthias Nie{\ss}ner, Matthew Fisher, Jianxiong Xiao, and Thomas Funkhouser.
\newblock 3dmatch: Learning local geometric descriptors from rgb-d reconstructions.
\newblock In \emph{Proceedings of the IEEE conference on computer vision and pattern recognition}, pages 1802--1811, 2017.

\bibitem[Zhao et~al.(2020)Zhao, Wu, Wang, Lin, Xia, Shen, Li, and Consortium]{zhao2020unsupervised}
Fenqiang Zhao, Zhengwang Wu, Li Wang, Weili Lin, Shunren Xia, Dinggang Shen, Gang Li, and UNC/UMN Baby Connectome~Project Consortium.
\newblock Unsupervised learning for spherical surface registration.
\newblock In \emph{Machine Learning in Medical Imaging: 11th International Workshop, MLMI 2020, Held in Conjunction with MICCAI 2020, Lima, Peru, October 4, 2020, Proceedings 11}, pages 373--383. Springer, 2020.

\bibitem[Zhao et~al.(2021{\natexlab{a}})Zhao, Wu, Wang, Lin, Xia, Shen, Wang, and Li]{zhao2021s3reg}
Fenqiang Zhao, Zhengwang Wu, Fan Wang, Weili Lin, Shunren Xia, Dinggang Shen, Li Wang, and Gang Li.
\newblock S3reg: superfast spherical surface registration based on deep learning.
\newblock \emph{IEEE transactions on medical imaging}, 40\penalty0 (8):\penalty0 1964--1976, 2021{\natexlab{a}}.

\bibitem[Zhao et~al.(2021{\natexlab{b}})Zhao, Liang, Wang, and Yang]{zhao2021centroidreg}
Hengwang Zhao, Zhidong Liang, Chunxiang Wang, and Ming Yang.
\newblock Centroidreg: A global-to-local framework for partial point cloud registration.
\newblock \emph{IEEE Robotics and Automation Letters}, 6\penalty0 (2):\penalty0 2533--2540, 2021{\natexlab{b}}.

\end{thebibliography}
